\documentclass[10pt,journal,compsoc]{IEEEtran}
\usepackage{amsmath,amsfonts}
\usepackage{algorithmic}
\usepackage{algorithm}
\usepackage{array}
\usepackage[caption=false,font=normalsize,labelfont=sf,textfont=sf]{subfig}
\usepackage{textcomp}
\usepackage{stfloats}
\usepackage{url}
\usepackage{verbatim}
\usepackage{graphicx}
\usepackage{cite}
\usepackage{graphicx}
\newcommand{\etal}{\textit{et al}. }
\newcommand{\ie}{\textit{i}.\textit{e}.}

\usepackage{makecell}
\usepackage{multirow}
\usepackage{booktabs}

\usepackage{adjustbox}
\usepackage{color,xcolor,colortbl}
\usepackage{bbding}
\usepackage{balance}

\newcommand{\widthscalefive}{0.28}
\newcommand{\widthscalethree}{0.4}
\definecolor{lightgray}{gray}{0.95}
\definecolor{lightblue}{HTML}{95bddc}

\usepackage[pagebackref=true,breaklinks=true,colorlinks,bookmarks=false,citecolor=green ]{hyperref}

\begin{document}

\title{

DiffI2I: Efficient Diffusion Model for Image-to-Image Translation}

\author{Bin Xia,
        Yulun Zhang,
        Shiyin Wang,
        Yitong Wang,
        Xinglong Wu,
        Yapeng Tian, \\
        Wenming Yang, 
        Radu Timotfe,
        Luc Van Gool
        %~\IEEEmembership{Senior Member,~IEEE,}
        % ,~\IEEEmembership{Fellow,~IEEE}% <-this % stops a space
\IEEEcompsocitemizethanks{
\IEEEcompsocthanksitem Bin Xia and Wenming Yang are with the Shenzhen International Graduate School, Tsinghua University, Shenzhen 518055, China. Email: zjbinxia@gmail.com, 
yang.wenming@sz.tsinghua.edu.cn.
\IEEEcompsocthanksitem Yulun Zhang and Luc Van Gool are with the Computer Vision Lab, ETH Z\"{u}rich, Z\"{u}rich 8092, Switzerland. Email: yulun100@gmail.com, vangool@vision.ee.ethz.ch. 
\IEEEcompsocthanksitem Yapeng Tian is with the Department of Computer Science, The University of Texas at Dallas, Richardson, TX 75080, USA. E-Mail: yapeng.tian@utdallas.edu
\IEEEcompsocthanksitem Yitong Wang, Shiyin Wang, and Xinglong Wu are with ByteDance Inc, Shenzhen 518055, China. E-Mail: wangyitong@pku.edu.cn,
wangshiyin.ai@bytedance.com,
wuxinglong@bytedance.com.
\IEEEcompsocthanksitem Radu Timofte is with Computer Vision Lab, IFI $\&$ CAIDAS, University of W\"{u}rzburg, Germany. E-Mail: timofte.radu@uni-wuerzburg.de}% <-this % stops an unwanted space
} %\thanks{Manuscript received April 19, 2005; revised September 17, 2014.}}

\markboth{Journal of \LaTeX\ Class Files,~Vol.~14, No.~8, August~2021}%
{Shell \MakeLowercase{\textit{et al.}}: A Sample Article Using IEEEtran.cls for IEEE Journals}


\IEEEtitleabstractindextext{\begin{abstract}

The Diffusion Model (DM) has emerged as the SOTA approach for image synthesis. However, the existing DM cannot perform well on some image-to-image translation (I2I) tasks. Different from image synthesis, some I2I tasks, such as super-resolution, require generating results in accordance with GT images. Traditional DMs for image synthesis require extensive iterations and large denoising models to estimate entire images, which gives their strong generative ability but also leads to artifacts and inefficiency for I2I. To tackle this challenge, we propose a simple, efficient, and powerful DM framework for I2I, called DiffI2I. Specifically, DiffI2I comprises three key components: a compact I2I prior extraction network (CPEN), a dynamic I2I transformer (DI2Iformer), and a denoising network. We train DiffI2I in two stages: pretraining and DM training. For pretraining, GT and input images are fed into CPEN$_{S1}$ to capture a compact I2I prior representation (IPR) guiding DI2Iformer. In the second stage, the DM is trained to only use the input images to estimate the same IRP as CPEN$_{S1}$. Compared to traditional DMs, the compact IPR enables  DiffI2I to obtain more accurate outcomes and employ a lighter denoising network and fewer iterations. Through extensive experiments on various I2I tasks, we demonstrate that DiffI2I achieves SOTA performance while significantly reducing computational burdens.

\end{abstract}

\begin{IEEEkeywords}
Diffusion model,  image-to-image translation, image restoration, inpainting, super-resolution, motion deblurring, dense prediction.
\vspace{3mm}
\end{IEEEkeywords}}

\maketitle
\IEEEdisplaynontitleabstractindextext

% \fi
\IEEEpeerreviewmaketitle

\IEEEraisesectionheading{\section{Introduction}}
\IEEEPARstart{I}{mage}-to image translation (I2I) aims to learn a mapping function between two distinct image domains. This field has garnered significant attention due to its versatile application potential and has been implemented in various domains, including inpainting~\cite{LaMa, inpainting-GAN}, super-resolution~\cite{SRCNN, KDSR}, and semantic segmentation\cite{segformer, seaformer}. Currently, deep-learning-based I2I methods have achieved astonishing results by assimilating robust priors from extensive datasets.

Recently, Diffusion Models (DMs)~\cite{DDPM} have demonstrated competitive performance in image synthesis~\cite{DDPM2, score-diffusion, DDPM3, DDPM4} and I2I, including inpainting~\cite{repaint, LDM} and super-resolution~\cite{SR3, diffir, srdiff}. In particular, DMs are trained by iteratively denoising the image in a reverse diffusion process. They have demonstrated the efficacy of principled probabilistic diffusion modeling in achieving high-quality mapping from randomly sampled Gaussian noise to complex target distributions, such as realistic images or latent distributions. It is notable that DMs do not encounter mode-collapse or training instabilities commonly observed in GANs~\cite{LDM}.

However, traditional DMs designed for image synthesis perform diffusion and reverse processes on the whole images~\cite{DDPM} or feature maps~\cite{LDM}. This dense estimation framework can bring them a strong generation ability while also requiring a large number of iteration steps (approximately $50-1000$ steps) on large denoising models, which consumes massive computational resources. However, some I2I tasks, such as super-resolution and deblurring, are different from image synthesis tasks, which provide rich guidance information and need to generate results in accordance with ground-truth images (GT). Therefore, the traditional DM framework is not suitable for directly being applied to some I2I tasks, which not only tends to generate unpleasant artifacts but also leads to low efficiency and slow convergence.

In this paper, we aim to design a simple, effective, and extendable diffusion framework for image-to-image translation. To this end, we propose DiffI2I, which performs the diffusion and the reverse process on the compressed I2I prior representation (IPR) to guide I2I rather than on whole images. This can not only adequately use the mapping ability of DM but also improve efficiency and convergence speed for I2I. Specifically, since the transformer can model long-range pixel dependencies, we adopt the transformer structure to form Dynamic I2Iformer (DI2Iformer). 
 We train our DiffI2I in two stages: \textbf{(1)} In the first stage (Fig.~\ref{fig:method} (a)), we focus on developing a compact I2I prior extraction network (CPEN), which can extract compressed IPR from the ground-truth image. This compact IPR serves as guidance information for the DI2Iformer. Additionally, we introduce two crucial components for the DI2Iformer: the Dynamic Feed-Forward Network (DFFN) and Dynamic Attention (DA). Both DFFN and DA enable the DI2Iformer to effectively utilize the extracted IPR. Notably, CPEN and DI2Iformer are jointly optimized during this stage.
 \textbf{(2)}  In the second stage (Fig.~\ref{fig:method} (b)),  we train the DM to directly estimate the accurate IPR from the input images. As the IPR is compact and primarily contributes details for guiding the I2I, our DM can estimate accurate and effective results with only several iterations on a lightweight denoising network.

\begin{figure*}[t]
\scriptsize
\centering
\resizebox{1\linewidth}{!}{
\begin{tabular}{ccc}
    \hspace{-4mm} \includegraphics[height=0.22\textwidth]{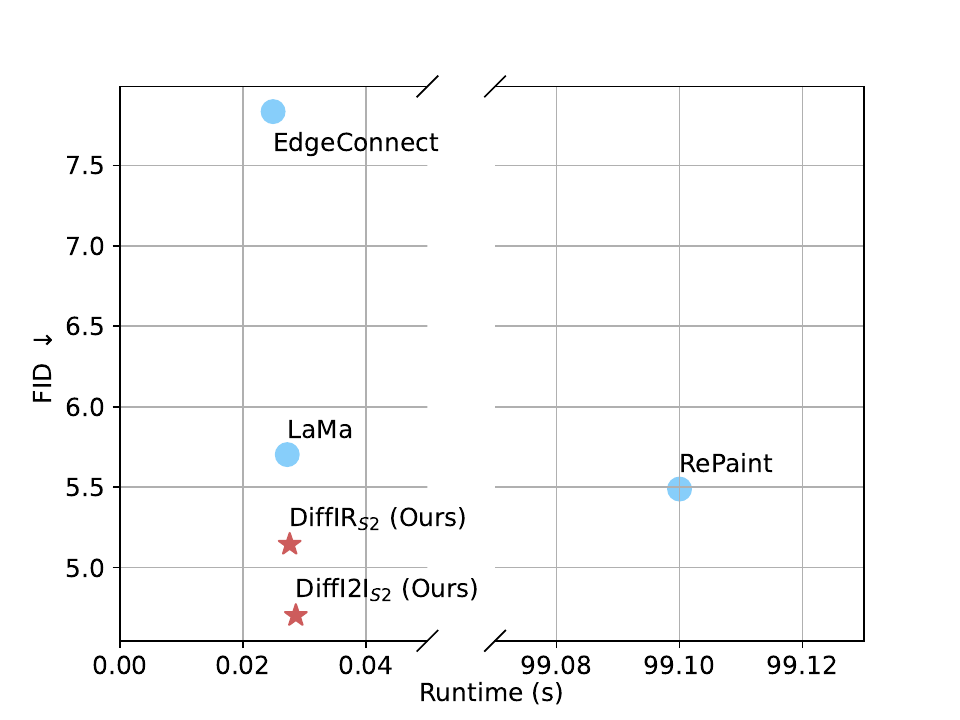}  & 
    \hspace{-6mm} \includegraphics[height=0.22\textwidth]{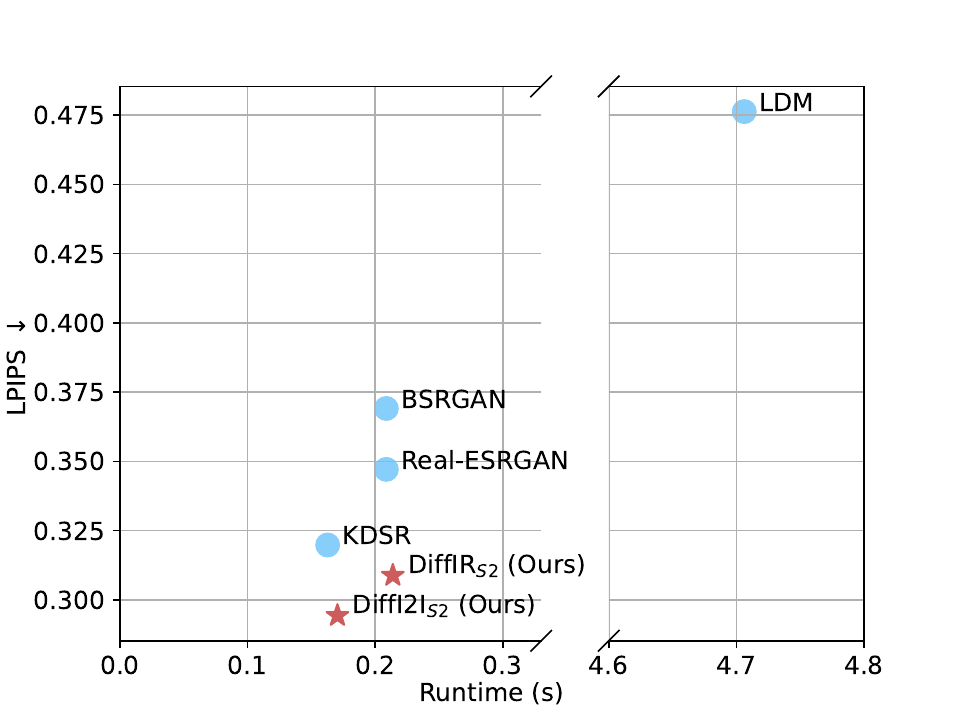}	 &
    \hspace{-6mm} \includegraphics[height=0.22\textwidth]{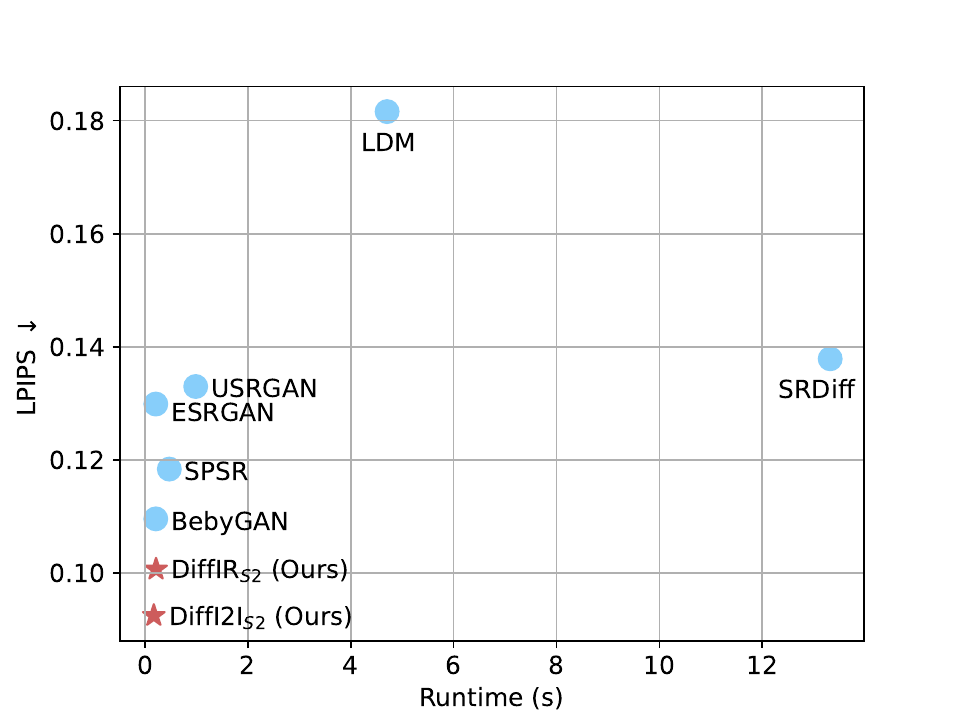}  \\

    \hspace{-5mm} \shortstack{(a) Inpainting (Tab.~\ref{tab:inpainting})}   &
    \hspace{-2mm} \shortstack{(b) Real-world super-resolution (Tab.~\ref{tab:real})}  &	
    \hspace{-2mm} \shortstack{(c) Single image super-resolution (Tab.~\ref{tab:SR})}  \\
    \hspace{-4mm} \includegraphics[height=0.22\textwidth]{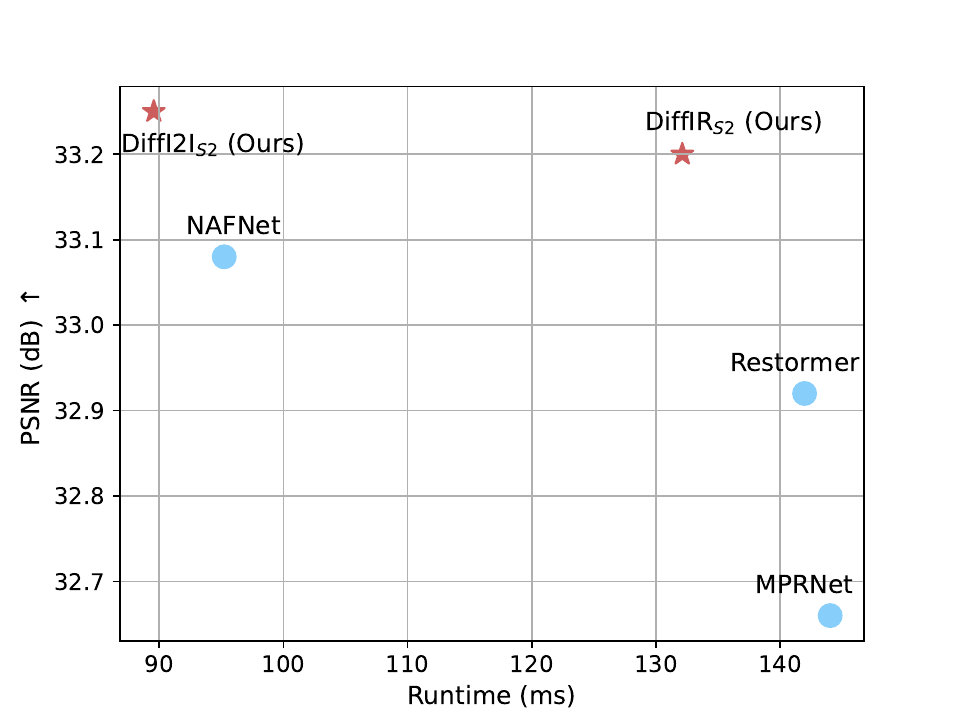}  & 
    \hspace{-6mm} \includegraphics[height=0.22\textwidth]{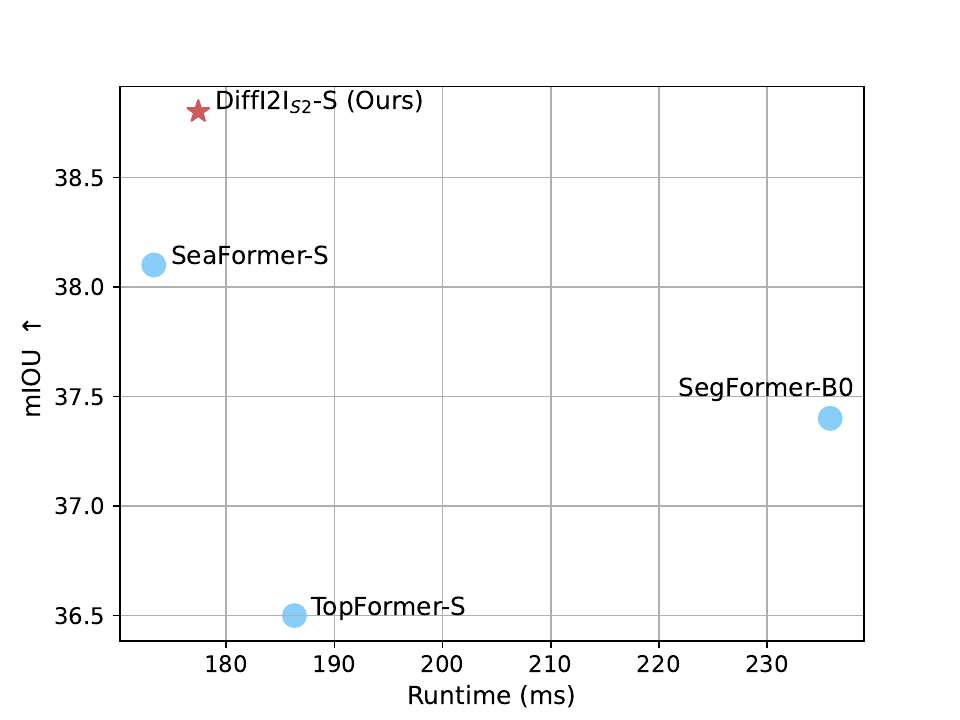}	 &
    \hspace{-6mm} \includegraphics[height=0.22\textwidth]{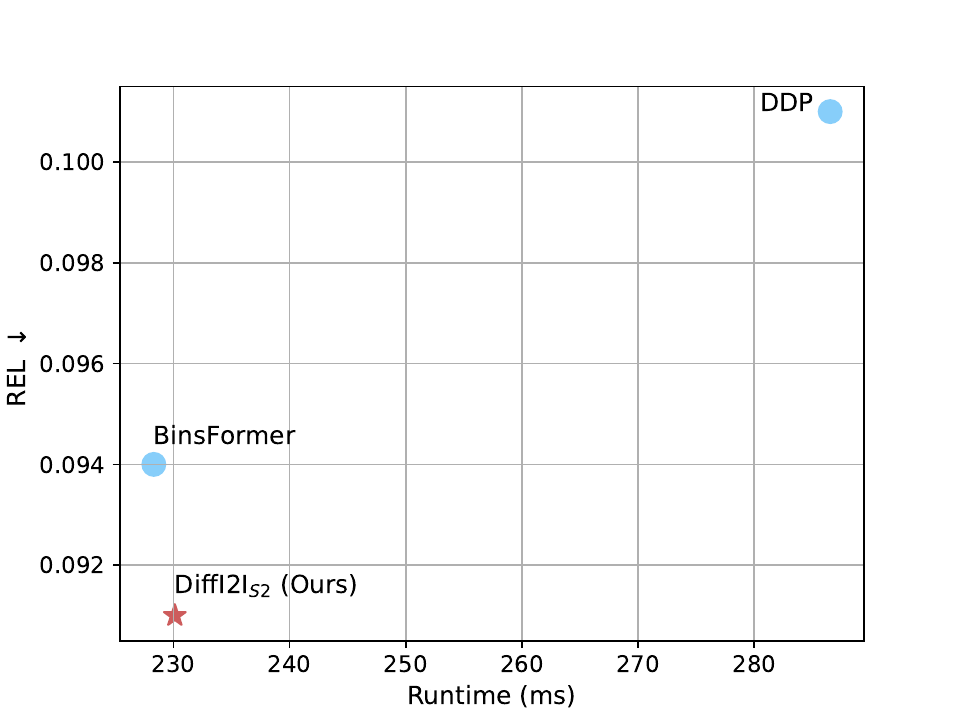}  \\

    \hspace{-5mm} \shortstack{(d) Motion deblurring (Tab.~\ref{tab:deblur})}   &
    \hspace{-2mm} \shortstack{(e) Semantic segmentation (Tab.~\ref{tab:seg})}   &	
    \hspace{-2mm} \shortstack{(f) Depth estimation (Tab.~\ref{tab:depth})}  
\end{tabular}
}
\caption{ 
Our DiffI2I achieves exceptional state-of-the-art performance in I2I tasks, all while maintaining an impressive level of efficiency. It is worth highlighting that existing methods like LDM~\cite{LDM}, RePaint~\cite{repaint}, SRDiff~\cite{srdiff}, and DDP~\cite{ddp} are all grounded in Differential Morphism (DM). Remarkably, DiffI2I stands out by achieving superior performance while boasting an astonishing efficiency improvement of \textbf{3500 times} over RePaint.
}
\label{fig:head}
\vspace{-2mm}
\end{figure*}

Furthermore, we develop a joint optimization of DM and decoder (\ie, DI2Iformer) scheme for DiffI2I. Specifically, traditional DM-based methods commonly separate the training of DM and decoder. This is because that traditional DM~\cite{LDM} consumes excessive computational costs and cannot run all iterations to optimize with the latter decoder jointly. However, a minor estimated error of DM would cause a performance drop of the decoder. Therefore, LDM~\cite{LDM} has to use the quantization dictionary to alleviate this problem. Fortunately, since the DM of our DiffI2I needs a few computational costs, we can conveniently execute all iterations and derive the estimated IPR to perform joint optimization with DI2Iformer. As illustrated in Fig.~\ref{fig:head}, DiffI2I surpasses the SOTA performance while significantly reducing runtime compared to other DM-based methods, such as RePaint~\cite{repaint}, SRDiff~\cite{srdiff}, and LDM~\cite{LDM}. Notably, DiffI2I achieves a remarkable efficiency improvement of 3500$\times$ over RePaint. Our contributions can be summarized into fourfold:
 \begin{itemize} 
 \vspace{0.5mm}
\item 
We introduce DiffI2I, a strong, simple, and efficient DM-based baseline for I2I. By leveraging powerful mapping capabilities, DiffI2I can estimate a compact IPR to guide the I2I process. This not only improves the performance of DM while enhancing efficiency, stability, and convergence speed on I2I.
 \vspace{0.5mm}
\item 
To fully harness the IPR for I2I, we propose Dynamic I2Iformer with DFFN and DA as essential components in the network structure.
 \vspace{0.5mm}
\item 
Unlike previous latent DMs that individually optimize denoising networks, we present joint optimization of the denoising network and DI2Iformer to further enhance the robustness of error estimation.
\item 
 \vspace{0.5mm}
Through extensive experiments, we demonstrate that DiffI2I achieves state-of-the-art (SOTA) performance in I2I tasks while significantly reducing runtime compared to other DM-based methods.
 \vspace{0.5mm}
\end{itemize}

A preliminary version of this work~\cite{diffir} has been accepted by ICCV 2023. In the current work, we introduce additional content in significant ways:
 \begin{itemize}
 \vspace{0.5mm}
\item 
We explore an efficient and effective DiffI2I framework, deploying it across various I2I tasks. These applications not only demonstrate the generality of our approach but also extend the potential breadth.
 \vspace{0.5mm}
\item 
We delve deeper into the particulars and enhance the initial version with extensive analyses, including assessments of resolution robustness, a comparison with the accelerated DM method, and examinations of the runtime.
 \vspace{0.5mm}
\item 
We extend our method for real-world super-resolution, semantic segmentation, and depth estimation.
Comprehensive benchmark experiments demonstrate that our method maintains its superiority over existing approaches in these I2I tasks.
\end{itemize}

\section{Related Work}

\subsection{Image-to-image Translation}
Image-to-image translation (I2I) has been widely studied in computer vision and has applications in  inpainting~\cite{LaMa,edgeconnect}, image super-resolution~\cite{SRCNN,EDSR}, image deblurring~\cite{MIMO-Unet,DBGAN}, semantic segmentation~\cite{FCN}, and depth estimation~\cite{va-depth}.  
Up to now, numerous CNN-based methods have emerged and achieved tremendous witnessed progress. Many of these advances focus on elaborate network architecture designs and learning schemes, such as residual block~\cite{VDSR,plug-denoiser,CAS-CNN}, GAN~\cite{WGAN-GP,ESRGAN,inpainting-GAN}, attention~\cite{RCAN,ENLCA,SAN,inpainting1,inpainting2,inpainting3,deepfillv2}, knowledge distillation~\cite{KDSR}, and multi-scale feature aggregation~\cite{deeplab,BTS,encoder-decoder}.

Recently, transformers, originally designed for natural language processing, have garnered significant attention in the computer vision community. In comparison to convolutional neural networks (CNN), transformers excel at capturing global interactions among various image regions, leading to state-of-the-art performance. As a result, transformers have found widespread adoption in numerous vision tasks, including image recognition~\cite{VIT,touvron2021training}, segmentation~\cite{seg1,segformer,topformer,seaformer,seg3}, object detection~\cite{detect1,detect2}, depth estimation~\cite{depthformer,binsformer}, and image restoration~\cite{IPT,SWINIR,restormer,MAT,NAFNet}.

\subsection{Diffusion Model}

Diffusion Models (DMs)~\cite{DDPM2, kingma2021variational, DDPM3} have emerged as a powerful approach in the domain of density estimation and sample quality, achieving impressive results across different modalities, such as images~\cite{LDM,diffir}, video~\cite{videodiff1,videodiff2}, audio~\cite{audiodiff1}, and biomedical~\cite{biodiff1,biodiff2}. In contrast to other generative models like GANs, DMs leverage parameterized Markov chains to optimize the lower variational bound on the likelihood function, empowering them to generate highly accurate target outputs. Considering the recent remarkable achievements of DMs in their respective domains, harnessing these models to develop I2I models presents a highly promising pathway to propel the boundaries of I2I tasks to new heights.

Recently, DMs have gained significant prominence in I2I tasks such as super-resolution~\cite{SR3}, inpainting~\cite{DDPM7, repaint,chung2022come,LDM}, semantic segmentation~\cite{Medsegdiff,wolleb2022diffusion}, and depth estimation~\cite{ddp,depthgen}. For example, SR3~\cite{SR3} and SRDiff~\cite{srdiff} introduced a DM for image super-resolution and demonstrated superior performance compared to traditional GAN-based methods. Palette~\cite{Palette} took inspiration from conditional generation models~\cite{mirza2014conditional} and proposed a conditional diffusion model for image restoration. Similarly, LDM~\cite{LDM} proposed a novel approach by applying DM on the latent space to enhance I2I efficiency, while RePaint designed an improved denoising strategy by resampling iterations in DM for inpainting. Moreover, Wolleb~\etal~\cite{wolleb2022diffusion} exploits the DM for medical segmentation, and DDP~\cite{ddp} extends
the denoising diffusion process into the dense prediction.

Nonetheless, all the aforementioned DMs follow the framework of traditional DM in image synthesis, which diffuse on the whole images or feature maps. Some I2I tasks are different from image synthesis, which pose a strong constraint and need to produce results in accordance with GT. Thus, the strong generative ability of the traditional DM not only tends to generate artifacts but also be inefficient and unstable. In this paper, we develop a simple, effective, and efficient DM framework for I2I, which extends DM on a compact IPR to improve performance and efficiency.

\section{Preliminaries: Diffusion Models}

A diffusion model (DM)~\cite{DDPM2} comprises two essential processes: the forward process (known as the diffusion process), and the reverse inference process.

During the training phase, DM methods define a diffusion process with a fixed Markov chain that converts an input image  $x_0$ into Gaussian noise $x_T \sim \mathcal{N}(0,1)$ by $T$ iterations. Each iteration of the diffusion process can be formulated as follows:
\begin{equation}
q\left(x_t \mid x_{t-1}\right)=\mathcal{N}\left(x_t ; \sqrt{1-\beta_t} x_{t-1}, \beta_t \mathbf{I}\right),
\label{eq:diff1}
\end{equation}
where $x_t$ is the noised image at time-step $t$, $\beta_t$ is the predefined scale factor, and $\mathcal{N}$ represents the Gaussian distribution. The Eq.~\eqref{eq:diff1} can be further simplified as follows:
\begin{equation}
q\left(\mathbf{x}_t \mid \mathbf{x}_0\right)=\mathcal{N}\left(\mathbf{x}_t ; \sqrt{\bar{\alpha}_t} \mathbf{x}_0,\left(1-\bar{\alpha}_t\right) \mathbf{I}\right),
\label{eq:diff2}
\end{equation}
where $\alpha_t=1-\beta_t$, $\bar{\alpha}_t=\prod_{i=0}^t \alpha_i$.

 In the inference stage (reverse process), DM methods sample a Gaussian random noise map $x_T$ and then iteratively and progressively denoise $x_T$ until it converges to a high-quality output $x_0$.
\begin{equation}
p\left(\mathbf{x}_{t-1} \mid \mathbf{x}_t, \mathbf{x}_0\right)=\mathcal{N}\left(\mathbf{x}_{t-1} ; \boldsymbol{\mu}_t\left(\mathbf{x}_t, \mathbf{x}_0\right), \sigma_t^2 \mathbf{I}\right),
\label{eq:diff3}
\end{equation}
where the mean $\boldsymbol{\mu}_t\left(\mathbf{x}_t, \mathbf{x}_0\right)$ is defined as $\boldsymbol{\mu}_t\left(\mathbf{x}_t, \mathbf{x}_0\right)=\frac{1}{\sqrt{\alpha_t}}\left(\mathbf{x}_t-\epsilon \frac{1-\alpha_t}{\sqrt{1-\bar{\alpha}_t}}\right)$ and the $\epsilon$ represents the noise in $x_t$, which is the only uncertain variable in the reverse process. The variance $\sigma_t$ is defined as $\sigma_t^2=\frac{1-\bar{\alpha}_{t-1}}{1-\bar{\alpha}_t} \beta_t$. To estimate $\epsilon$, DMs employ a denoising network $\epsilon_{\theta}(x_t,t)$. To train $\epsilon_{\theta}(x_t,t)$, given a clean image $x_0$, DMs randomly sample a time step $t$ and a noise $\epsilon \sim \mathcal{N}(0, \mathbf{I})$ to generate noisy images $x_t$ using Eq.~\eqref{eq:diff2}. Subsequently, DMs optimize the network parameters $\theta$ of $\epsilon_{\theta}$ following~\cite{DDPM2}:
\begin{equation}
\nabla_{\boldsymbol{\theta}}\left\|\epsilon-\epsilon_{\boldsymbol{\theta}}\left(\sqrt{\bar{\alpha}_t} \mathrm{x}_0+\epsilon \sqrt{1-\bar{\alpha}_t}, t\right)\right\|_2^2.
\label{eq:diff4}
\end{equation}

\begin{figure*}[t]
	\centering
  \resizebox{1\linewidth}{!}{
	\includegraphics[height=10cm]{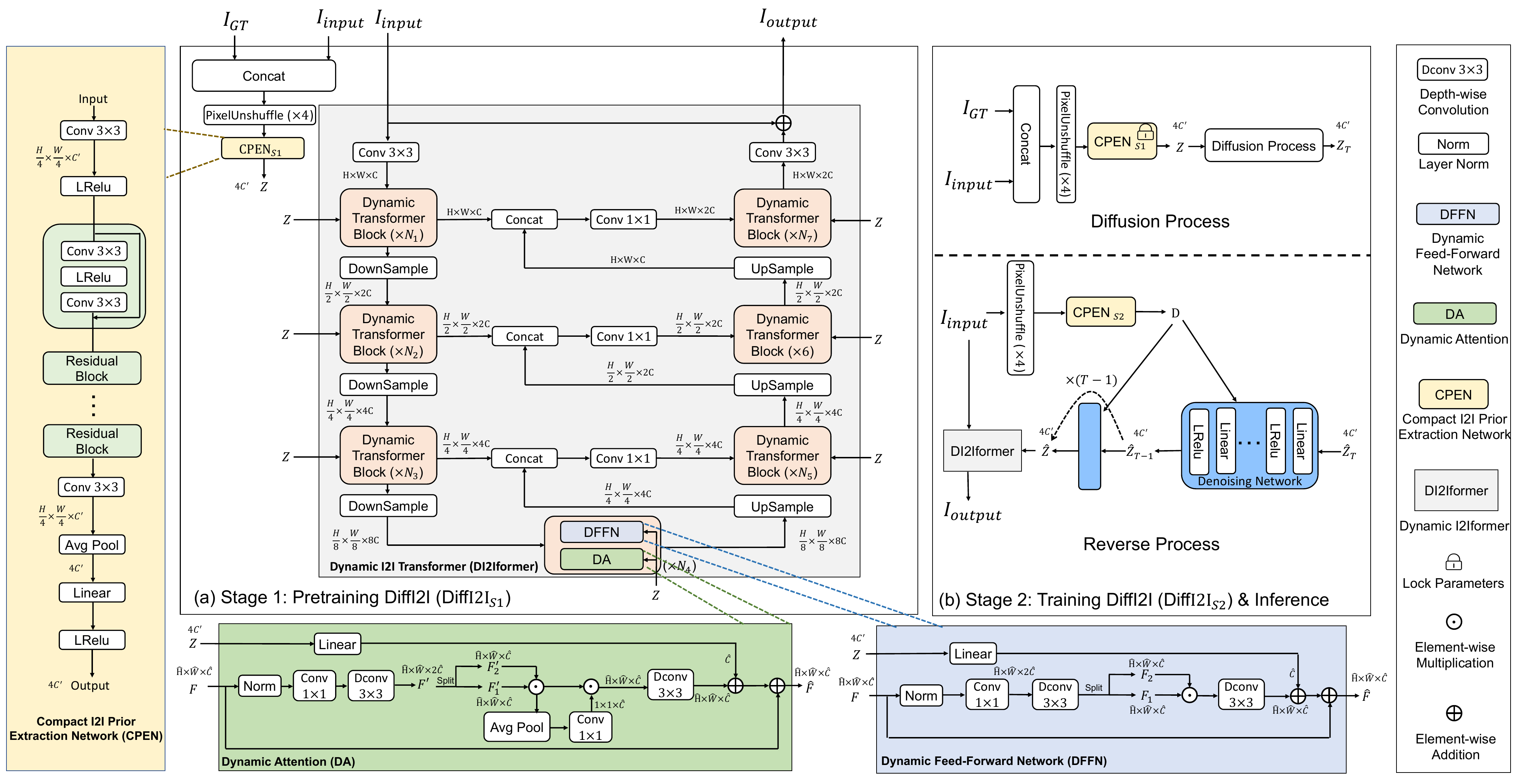}
 }
 \vspace{-5mm}
	\caption{The overview of the proposed DiffI2I, which consists of three main components: DI2Iformer, CPEN, and a denoising network. DiffI2I has two training stages: (a) In the first stage, CPEN$_{S1}$ takes the GT image as input and produces an IPR $\mathbf{Z}$. This IPR guides DI2Iformer to finish I2I. The training of CPEN$_{S1}$ is optimized jointly with DI2Iformer to enable DI2Iformer to effectively utilize the extracted IPR. (b) Moving on to the second stage, we use the strong data estimation abilities of the DM to estimate the IPR extracted by pretrained CPEN$_{S1}$. Notably, we do not provide the GT image as input to CPEN$_{S2}$ and the denoising network. During the inference stage, we employ the reverse process of DM for I2I.  }
	\label{fig:method}
 \vspace{-1mm}
\end{figure*}

\begin{algorithm*}[t]
	\caption{ DiffI2I$_{S2}$ Training}
	\label{alg:train_DiffI2I}
	\textbf{Input}: Trained DiffI2I$_{S1}$ (including CPEN$_{S1}$ and DI2Iformer), $\beta_t (t\in[1,T])$. \\
	\textbf{Output}: Trained DiffI2I$_{S2}$. \\
        \vspace{-4mm}
	\begin{algorithmic}[1] %[1] enables line numbers
		\STATE Init: $\alpha_t=1-\beta_t$, $\bar{\alpha}_T=\prod_{i=0}^T \alpha_i$.
            \STATE Init: The DI2Iformer of DiffI2I$_{S2}$ copies the parameters of trained DiffI2I$_{S1}$. 
            \FOR{$I_{Input}$,  $I_{GT}$ }
            \STATE    $\mathbf{Z}=\operatorname{CPEN_{S1}}(\operatorname{PixelUnshuffle}(\operatorname{Concat}(I_{GT},I_{Input}))). $ ( Eq.~\eqref{eq:cpen})
            \STATE \textbf{Diffusion Process}:
            \STATE  We sample $\mathbf{Z}_{T}$ by $q\left(\mathbf{Z}_T \mid               \mathbf{Z}\right)=\mathcal{N}\left(\mathbf{Z}_T; \sqrt{\bar{\alpha}_T} \mathbf{Z},\left(1-\bar{\alpha}_T\right) \mathbf{I}\right)$  (\ie, diffusion process.  Eq.~\eqref{eq:mydiff1}) 
            \STATE \textbf{Reverse Process}:
            \STATE $\mathbf{\hat{Z}}_T = \mathbf{Z}_{T}$
            \STATE $\mathbf{D}=\operatorname{CPEN_{S2}}(\operatorname{PixelUnshuffle}(I_{Input}))$ (Eq.~\eqref{eq:cpen2})
            \FOR{$t=T$ to $1$ }
            \STATE  $\mathbf{\hat{Z}}_{t-1}=\frac{1}{\sqrt{\alpha_t}}\left(\mathbf{\hat{Z}}_t-\epsilon_{\theta}(\operatorname{Concat}(\mathbf{\hat{Z}}_t,t,\mathbf{D})) \frac{1-\alpha_t}{\sqrt{1-\bar{\alpha}_t}}\right)$ ( Eq.~\eqref{eq:mydiff2})               
            \ENDFOR
            \STATE $\mathbf{\hat{Z}}=\mathbf{\hat{Z}}_{0}$
            \STATE $I_{Output} = \operatorname{DI2Iformer}(I_{Input},\mathbf{\hat{Z}})$
            \STATE Calculate $\mathcal{L}_{diff}$ loss ( Eq.~\eqref{eq:diff}).
            \ENDFOR     
		\STATE Output the trained model DiffI2I$_{S2}$.
	\end{algorithmic}
 \label{alg:train}
\end{algorithm*}

\begin{algorithm*}[t]
	\caption{ DiffI2I$_{S2}$ Inference}
	\label{alg:infer_DiffI2I}
	\textbf{Input}: Trained DiffI2I$_{S2}$ (including CPEN$_{S2}$ and DI2Iformer), $\beta_t (t\in[1,T])$, input images $I_{Input}$. \\
	\textbf{Output}: Output images $\hat{I}_{Output}$. \\
        \vspace{-4mm}
	\begin{algorithmic}[1] %[1] enables line numbers
		\STATE Init: $\alpha_t=1-\beta_t$, $\bar{\alpha}_T=\prod_{i=0}^T \alpha_i$.
            \STATE \textbf{Reverse Process}:
            \STATE Sample $\mathbf{\hat{Z}}_T \sim \mathcal{N}(0,1)$
            \STATE $\mathbf{D}=\operatorname{CPEN_{S2}}(\operatorname{PixelUnshuffle}(I_{Input}))$ ( Eq.~\eqref{eq:cpen2})
            \FOR{$t=T$ to $1$ }
            \STATE  $\mathbf{\hat{Z}}_{t-1}=\frac{1}{\sqrt{\alpha_t}}\left(\mathbf{\hat{Z}}_t-\epsilon_{\theta}(\operatorname{Concat}(\mathbf{\hat{Z}}_t,t,\mathbf{D})) \frac{1-\alpha_t}{\sqrt{1-\bar{\alpha}_t}}\right)$ ( Eq.~\eqref{eq:mydiff2})               
            \ENDFOR
            \STATE $\mathbf{\hat{Z}}=\mathbf{\hat{Z}}_{0}$
            \STATE $I_{Output} = \operatorname{DI2Iformer}(I_{Input},\mathbf{\hat{Z}})$
   
		\STATE Output images $I_{Output}$.
	\end{algorithmic}
 \label{alg:infer}
\end{algorithm*}

\section{Methodology}

Recently, some DM-based methods~\cite{srdiff,SR3,repaint,ddp} follow the DM framework of image synthesis.
However, Some I2I tasks have strong constraints to generate results in accordance with GT, while image synthesis does not. Considering the difference between I2I and image synthesis, it is necessary to design a DM framework for I2I in particular. Traditional DMs perform the diffusion process on entire images or feature maps, which tend to generate artifacts and require a large denoising network and quite a few iterations.   To alleviate the issue, we introduce an effective and efficient approach for I2I, called DiffI2I, which utilizes DM to estimate a compressed IPR. The IPR serves as guidance information for the DI2Iformer to generate outputs. This can not only effectively leverage the mapping ability of DM to produce precise outputs but also necessitate a less complex denoising network and fewer iterations.

In this section, we take the image restoration task as an example to introduce  DiffI2I, as illustrated in Fig.~\ref{fig:method}. For other tasks, such as depth estimation, we incorporate IPR into the  SWIN Transformer~\cite{swintranformer} in a similar manner. DiffI2I is primarily composed of three components: a compact IR prior extraction network (CPEN), a dynamic I2Iformer (DI2Iformer), and a denoising network. The training of DiffI2I involves two stages: pretraining DiffI2I and training the diffusion model. In the following sections, we will first discuss the pretraining of DiffI2I in Sec.~\ref{sec:pretrain}, followed by the details of efficiently training the diffusion model for DiffI2I in Sec.~\ref{sec:diffusion}.

\subsection{Pretrain DiffI2I}
\label{sec:pretrain}
Before introducing pretraining DiffI2I, we would like to introduce two networks in the first stage: the compact I2I prior extraction network (CPEN) and the dynamic I2Iformer (DI2Iformer).  The architecture of CPEN is depicted in Fig.~\ref{fig:method} (yellow box), and it primarily consists of stacked residual blocks and linear layers to extract the compact I2I Prior Representation (IPR). Subsequently, the extracted IPR is utilized by DI2Iformer to translate images. 
The structure of DI2Iformer is represented in Fig.~\ref{fig:method} (pink box), which is composed of dynamic transformer blocks arranged in a Unet shape. These dynamic transformer blocks consist of two main components: dynamic  attention (DA) shown in Fig.~\ref{fig:method} (green box) and dynamic feed-forward network (DFFN) illustrated in Fig.~\ref{fig:method} (nattier blue box). These components leverage the extracted IPR as dynamic modulation parameters to incorporate I2I guidance information into feature maps effectively.

% compressed I2I prior representation
In order to solve the problem of artifacts and inefficiency caused by traditional DM performing diffusion on the entire images, we conducted diffusion on a compressed I2I prior representation (IPR). In the pretraining stage, we train CPEN$_{S1} $ to learn an effective and compact IPR to guide DI2Iformer.
To this end, as shown in Fig.~\ref{fig:method}~(a), we simultaneously train CPEN$_{S1}$ and DI2Iformer together. To begin, we concatenate the ground-truth (GT) and input images and apply the PixelUnshuffle operation to downsample them, resulting in the input for CPEN$_{S1}$. Next, CPEN$_{S1}$ extracts the IPR $\mathbf{Z}\in \mathbb{R}^{4C^{\prime}}$ using the following process:
\begin{equation}
\label{eq:cpen}
\mathbf{Z}=\operatorname{CPEN_{S1}}(\operatorname{PixelUnshuffle}(\operatorname{Concat}(I_{GT},I_{Input}))). 
\end{equation}
By concatenating GT and input images as input to CPEN$_{S1}$, and then training CPEN$_{S1}$ and DI2Iformer collaboratively, CPEN$_{S1}$ becomes proficient in deriving an effective IPR. This learned IPR effectively steers the DI2Iformer in producing accurate outputs.

For the DA block of DI2Iformer (Fig.~\ref{fig:method}~(a)), we use channel attention to efficiently aggregate global information. Specifically, given feature maps $\mathbf{F}\in\mathbb{R}^{\hat{H} \times \hat{W} \times \hat{C}}$, it is projected into feature maps $\mathbf{F^{\prime}}=W_d W_c \operatorname{Norm}(\mathbf{F})$, where
$W_c$ is the $1\times1$ point-wise convolution,  $W_d$ is the $3\times3$ depth-wise convolution, and $\operatorname{Norm}$ denotes layer normalization~\cite{layernorm}. Then, we use the SimpleGate ($\operatorname{SG}(.)$) as the activation function, which is the simplified version of GELU~\cite{GELU} and can be described as Eq.~\eqref{eq:sg}. After that, we use a global average pooling operation to aggregate the spatial information into channels and multiply global average pooling results with the input to weight the feature map (Eq.~\eqref{eq:CA}). 
\begin{equation}
\label{eq:sg}
\mathbf{F^{\prime}}_{SG}=\operatorname{SG}(\mathbf{F^{\prime}}_{1},\mathbf{F}^{\prime}_{2})=\mathbf{F}^{\prime}_{1} \odot \mathbf{F^{\prime}}_{2},
\end{equation}
\begin{equation}
\label{eq:CA}
\mathbf{F}^{\prime}_{CA}=\mathbf{F^{\prime}}_{SG}\odot\phi(\mathbf{F^{\prime}}_{SG}),
\end{equation}
where  $\mathbf{F^{\prime}_{1}},\mathbf{F^{\prime}_{2}}\in\mathbb{R}^{\hat{H} \times \hat{W} \times \hat{C}}$ are obtained by equally splitting $\mathbf{F^{\prime}}$. $\phi$ indicates the global average pooling operation. $\odot$ indicates element-wise multiplication. $\mathbf{F^{\prime}}_{SG},\mathbf{F^{\prime}}_{CA}\in\mathbb{R}^{\hat{H} \times \hat{W} \times \hat{C}}$ and $\phi(\mathbf{F^{\prime}}_{SG})\in\mathbb{R}^{ \hat{C}}$.

% Subsequently, we add the IPR into DA for guidance. The overall process of DA can be described as Eq.~\eqref{eq:DA}.

Then, we use the extracted IPR $\mathbf{Z}$ as dynamic modulation parameters for the DA of the DI2Iformer to guide I2I.  The overall process of DA can be described as follows:
\begin{equation}
\hat{\mathbf{F}}=W_d\mathbf{F}^{\prime}_{CA} + W_{l}\mathbf{Z}+\mathbf{F},
\end{equation}
where  $W_{l}$ represents linear layer, $\mathbf{F}$ and $\hat{\mathbf{F}} \in\mathbb{R}^{\hat{H}\times\hat{W}\times\hat{C}}$ are input and output feature maps respectively, and $W_{l}\mathbf{Z}\in\mathbb{R}^{\hat{C}}$.

Next, in DFFN, we aggregate local features. To achieve this, we utilize a $1\times1$ Conv layer to aggregate information from different channels and a $3\times3$ depth-wise Conv layer to aggregate information from spatially neighboring pixels. Additionally, we also use the SimpleGate (Eq.~\eqref{eq:sg}) as the activation function as DA does. Similar to DA, we incorporate IPR into DFFN to guide I2I. The overall process of DFFN can be defined as:
\begin{equation}
\mathbf{\hat{F}}=W_d^2\operatorname{SG}\left(W_d^1 W_c\operatorname{Norm}(\mathbf{F})\right) +W_{l}\mathbf{Z}+\mathbf{F}.
\end{equation}
We jointly train CPEN$_{S1}$ and DI2Iformer to enable integration of IPR extracted by CPEN$_{S1}$ into the I2I process performed by DI2Iformer. This empowers DI2Iformer to leverage the guidance information from CPEN$_{S1}$ for enhancing the quality and accuracy of the I2I.

\subsection{Diffusion Models for I2I}
\label{sec:diffusion}

The training and inference procedures of DiffI2I$_{S2}$ (Fig.~\ref{fig:method}~(b)) are illustrated in Algs.~\ref{alg:train} and \ref{alg:infer}, respectively. Detailed explanations of these procedures are given in the subsequent parts.

In the second stage (Fig.~\ref{fig:method}~(b)), we capitalize on the powerful data estimation capabilities of the DM to estimate IPR without inputting GT images. To achieve this, we leverage the pre-trained CPEN$_{S1}$ to capture the IPR $\mathbf{Z}\in \mathbb{R}^{4C^{\prime}}$. Subsequently, we subject $\mathbf{Z}$ to a diffusion process to obtain a sample $\mathbf{Z}_T\in \mathbb{R}^{4C^{\prime}}$. This diffusion process can be described as follows:
\begin{equation}
q\left(\mathbf{Z}_T \mid \mathbf{Z}\right)=\mathcal{N}\left(\mathbf{Z}_T; \sqrt{\bar{\alpha}_T} \mathbf{Z},\left(1-\bar{\alpha}_T\right) \mathbf{I}\right),
\label{eq:mydiff1}
\end{equation}
where $T$ represents the total number of iterations, $\bar{\alpha}$ and $\alpha$ are defined in Eqs.~\eqref{eq:diff1} and ~\eqref{eq:diff2} (\ie,  $\bar{\alpha}_T=\prod_{i=0}^T \alpha_i$).

In the reverse process, DiffI2I$_{S2}$ can achieve quite accurate estimations using much fewer iterations and a smaller model size compared to traditional DMs~\cite{LDM, repaint}, mainly because DiffI2I$_{S2}$ diffuses on compact IPR. 

Moreover, we introduce joint optimization of DMs and decoders in DiffI2I$_{S2}$, which can further improve performance. Traditional DMs incur significant computational costs due to the large  denoising network and the number of iterations they require. Consequently, they are forced to randomly sample a time-step $t\in[1,T]$ and optimize the denoising network solely at that time step (Eqs.~\eqref{eq:diff1}, \eqref{eq:diff2}, \eqref{eq:diff3}, and~\eqref{eq:diff4}). Unfortunately, this lack of joint training between the denoising network and the decoder (\ie, DI2Iformer) means that minor errors in estimations from the denoising network would make the performance drop of DI2Iformer.
In contrast, the lightweight DM framework employed in DiffI2I$_{S2}$ offers a notable advantage. It initiates its iterations from the $T$-th time step (Eq.~\eqref{eq:mydiff1}), facilitating the execution of all denoising iterations (Eq.~\eqref{eq:mydiff2}) to derive $\mathbf{\hat{Z}}$. After that, $\mathbf{\hat{Z}}$ can be seamlessly integrated into the joint optimization process with DI2Iformer.

\begin{equation}
\mathbf{\hat{Z}}_{t-1}=\frac{1}{\sqrt{\alpha_t}}\left(\mathbf{\hat{Z}}_t-\epsilon \frac{1-\alpha_t}{\sqrt{1-\bar{\alpha}_t}}\right),
\label{eq:mydiff2}
\end{equation}
where $\epsilon$ indicates the same noise settings in Eq.~\eqref{eq:diff3}. We employ CPEN$_{S2}$ and the denoising network to predict the noise using Eq.~\eqref{eq:diff3}. Notably, different from traditional DMs in Eq.~\eqref{eq:diff3}, our DiffI2I$_{S2}$ eliminates the variance estimation. This modification proves beneficial, leading to more accurate IPR estimation and improved overall performance (as discussed in Sec.~\ref{sec:ablation}).

Specifically, in the reverse process of DM, we begin by using CPEN$_{S2}$ to derive a conditional vector $\mathbf{D} \in\mathbb{R}^{4C^{\prime}}$ from input images:
\begin{equation}
\label{eq:cpen2}
\mathbf{D}=\operatorname{CPEN_{S2}}(\operatorname{PixelUnshuffle}(I_{input})),
\end{equation}
% where CPEN$_{S2}$ has the same structure as CPEN$_{S1}$ except the input dimension of the first convolution. Then, we use the denoising network $\epsilon_{\theta}$ to estimate noise in each time step $t$ as  $\epsilon_{\theta}(\operatorname{Concat}(\mathbf{\hat{Z}}_t,t,\mathbf{D}))$. The estimated noise is substituted into Eq.~\eqref{eq:mydiff2} to obtain $\mathbf{\hat{Z}}_{t-1}$ to start the next iteration. 
where CPEN$_{S2}$ shares the same structure as CPEN$_{S1}$, with the only difference being the input dimension of the first convolution. Subsequently, we utilize the denoising network $\epsilon_{\theta}$ to estimate the noise for each time step $t$, which is computed as $\epsilon_{\theta}(\operatorname{Concat}(\mathbf{\hat{Z}}_t,t,\mathbf{D}))$. The estimated noise is then utilized in Eq.~\eqref{eq:mydiff2} to calculate $\mathbf{\hat{Z}}_{t-1}$, which serves as the starting point for the next iteration.

\begin{table*}[t]
  \centering
  \caption{Quantitative evaluation (FID/LPIPS) for \textbf{inpainting} on benchmark datasets. The best and second-best performances are highlighted in bold and underlined, respectively. Notably, the last four methods, shaded in gray, utilize the diffusion model.}
  \resizebox{1\linewidth}{!}{
    \begin{tabular}{l|c|cccc|cccc}
    \toprule[0.2em]
    \multirow{3}[5]{*}{\textbf{Method}} & \multirow{3}[5]{*}{\textbf{\#Params (M)}} & \multicolumn{4}{c|}{\textbf{Places~\cite{places2} (512$\times$512)}} & \multicolumn{4}{c}{\textbf{CelebA-HQ~\cite{celeba} (256$\times$256)}} \\
\cmidrule{3-10}          &       & \multicolumn{2}{c}{Narrow Masks} & \multicolumn{2}{c|}{Wide Masks} & \multicolumn{2}{c}{Narrow Masks} & \multicolumn{2}{c}{Wide Masks} \\
\cmidrule{3-10}          &       & FID $\downarrow$  & LPIPS $\downarrow$ & FID $\downarrow$  & LPIPS $\downarrow$& FID  $\downarrow$ & LPIPS $\downarrow$& FID $\downarrow$  & LPIPS $\downarrow$\\
\midrule
    EdgeConnect~\cite{edgeconnect} & 22    & 1.3421 & 0.1106 & 8.4866 & 0.1594 & 6.9566 & 0.0922 & 7.8346 & 0.1149 \\
    ICT~\cite{ICT}   & 150   & -     & -     & -     & -     &   8.4977	& 0.0982 & 9.8794 &	0.1196\\
    % LaMa-big & 51    & 0.5116 & 0.0808 & 1.8522 & 0.1286 & -     & -     & -     & - \\
    LaMa~\cite{LaMa}  & 27    & 0.6340 & 0.0898 & 2.2494 & 0.1339 & 5.3889 & 0.0806 & 5.7023 & 0.0951 \\
    \rowcolor{lightgray}
    LDM~\cite{LDM}   & 215   & -     & -     & 2.1500  & 0.1440 & -     & -     & -     & - \\
    \rowcolor{lightgray}
    RePaint~\cite{repaint} & 607   & -     & -     & -     & -     &  4.7395	& 0.0890	& 5.4881 & 0.1094  \\
    % DiffIRs2-big & 49    & 0.4702 & 0.0745 & 1.7584 & 0.1276 & -     & -     & -     & - \\
    \rowcolor{lightgray}
    DiffIR$_{S2}$~\cite{diffir} (Ours)  & 26    & \underline{0.4913} & \underline{0.0758} & \underline{1.9788} & \underline{0.1306} & \underline{4.5967} & \underline{0.0769} & \underline{5.1440} & \underline{0.0918} \\
    \rowcolor{lightgray}
    DiffI2I$_{S2}$ (Ours) &	27 &	\textbf{0.4452} &	\textbf{0.0706} & 	\textbf{1.7835} &	\textbf{0.1261} &	\textbf{3.8640} &	\textbf{0.0635} &	\textbf{4.7013} &	\textbf{0.0859} \\
    
    \bottomrule[0.2em]
    \end{tabular}%
    }
  \label{tab:inpainting}%
  \vspace{-1mm}
\end{table*}%

\begin{figure*}[t]
    \newlength\fsdurthree
    \setlength{\fsdurthree}{0mm}
    \Huge
    \centering
   \resizebox{1\linewidth}{!}{
            \begin{adjustbox}{valign=t}
                \begin{tabular}{cccccc}

                    \includegraphics[width=\widthscalethree  \textwidth]{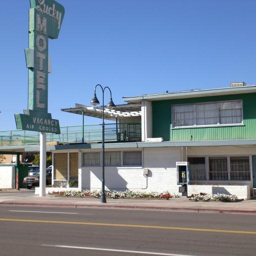} \hspace{\fsdurthree} &
                    \includegraphics[width=\widthscalethree  \textwidth]{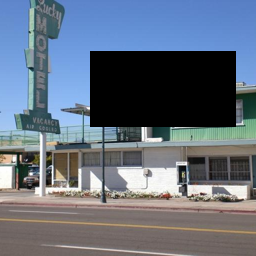} \hspace{\fsdurthree}  &
                    \includegraphics[width=\widthscalethree  \textwidth]{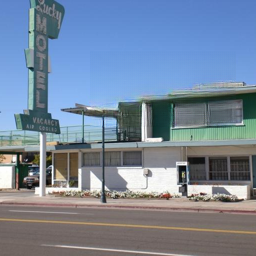} 
                    \hspace{\fsdurthree} &
                    \includegraphics[width=\widthscalethree  \textwidth]{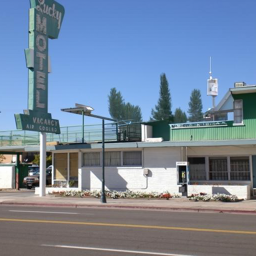} \hspace{\fsdurthree} &
                    \includegraphics[width=\widthscalethree  \textwidth]{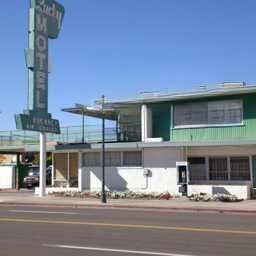} \hspace{\fsdurthree}  &
                    \includegraphics[width=\widthscalethree  \textwidth]{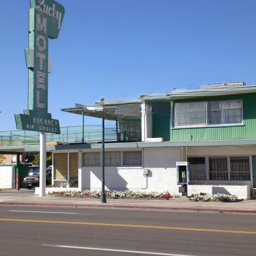} \hspace{\fsdurthree}
                    \\
                    \includegraphics[width=\widthscalethree  \textwidth]{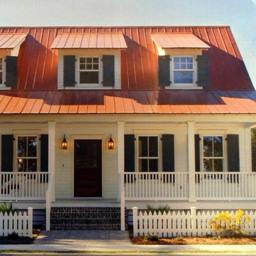} \hspace{\fsdurthree} &
                    \includegraphics[width=\widthscalethree  \textwidth]{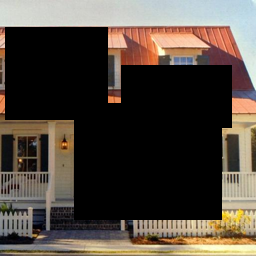} \hspace{\fsdurthree} &
                    \includegraphics[width=\widthscalethree  \textwidth]{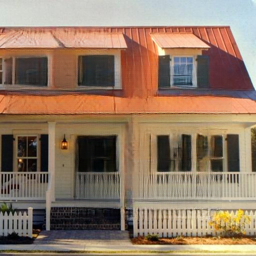} 
                    \hspace{\fsdurthree} &
                    \includegraphics[width=\widthscalethree  \textwidth]{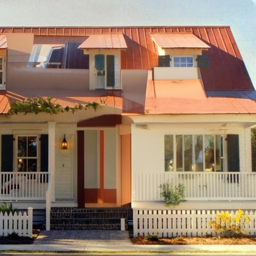} \hspace{\fsdurthree} &
                    \includegraphics[width=\widthscalethree  \textwidth]{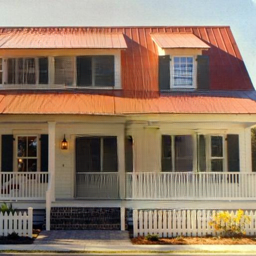} \hspace{\fsdurthree}  &
                    \includegraphics[width=\widthscalethree  \textwidth]{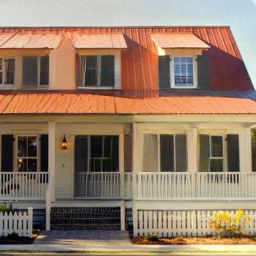} \hspace{\fsdurthree}
                    \\
                    \includegraphics[width=\widthscalethree \textwidth]{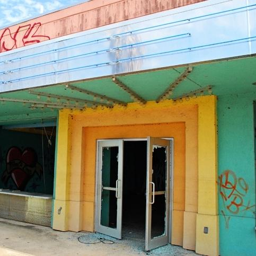} \hspace{\fsdurthree} &
                    \includegraphics[width=\widthscalethree  \textwidth]{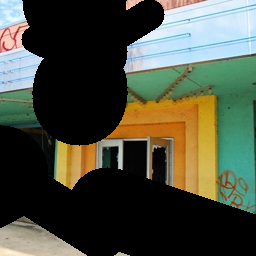} \hspace{\fsdurthree} &
                    \includegraphics[width=\widthscalethree  \textwidth]{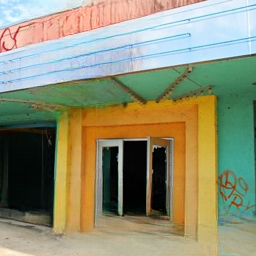} 
                    \hspace{\fsdurthree} &
                    \includegraphics[width=\widthscalethree  \textwidth]{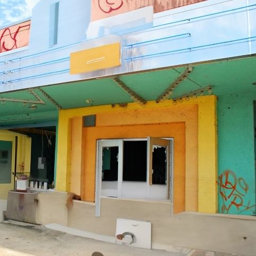} \hspace{\fsdurthree} &
                    \includegraphics[width=\widthscalethree  \textwidth]{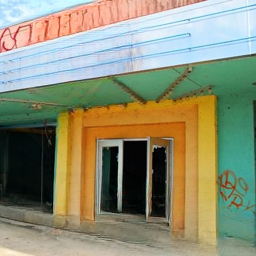} \hspace{\fsdurthree}  &             \includegraphics[width=\widthscalethree  \textwidth]{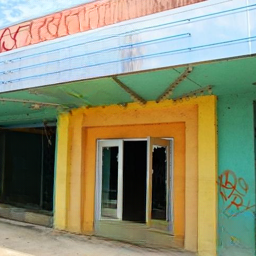} \hspace{\fsdurthree} 
                    \\
                    GT \hspace{\fsdurthree} &
                    \makecell{Input} \hspace{\fsdurthree} &
                    % \makecell{ICT~\cite{ICT}} \hspace{\fsdurthree} &
                    LaMa~\cite{LaMa} \hspace{\fsdurthree} &
                    RePaint~\cite{repaint} \hspace{\fsdurthree} &
                    \makecell{DiffIR$_{S2}$~\cite{diffir} (Ours)} \hspace{\fsdurthree} &
                     \makecell{DiffI2I$_{S2}$ (Ours)} \hspace{\fsdurthree} 
                \end{tabular}
            \end{adjustbox}

    }
    \caption{Visual comparison of \textbf{ inpainting} methods. Zoom-in for better details.}
    \label{fig:inpainting_show}
    \vspace{-2mm}
\end{figure*}

After performing $T$ iterations, we obtain the final estimated IPR $\mathbf{\hat{Z}} \in \mathbb{R}^{4C^{\prime}}$. We jointly train CPEN$_{S2}$, the denoising network, and DI2Iformer using the combined loss function $\mathcal{L}{all}$:
\begin{equation}
\label{eq:diff}
\mathcal{L}_{diff}=\frac{1}{4C^{\prime}}\sum_{i=1}^{4C^\prime}\left|\hat{\mathbf{Z}}(i)-\mathbf{Z}(i)\right |, 
\mathcal{L}_{all}=\mathcal{L}_{task}+\mathcal{L}_{diff},
\end{equation}
where $\mathcal{L}_{task}$ is the common task-specific loss. For example, we use $L_{rec}= \left\|I_{GT}-I_{Output}\right\|_{1}$ as $\mathcal{L}_{task}$ for deblurring. Moreover, we use the cross-entropy loss as $\mathcal{L}_{task}$ for semantic segmentation.

During the inference stage, only the reverse diffusion process is employed, as depicted in the lower part of Fig.~\ref{fig:method}~(b). CPEN$_{S2}$ is responsible for extracting a conditional vector $\mathbf{D}$ from the input images, and a Gaussian noise  $\mathbf{\hat{Z}}_T$ is randomly sampled. The denoising network utilizes both $\mathbf{\hat{Z}}_T$ and $\mathbf{D}$ to estimate the IPR $\mathbf{\hat{Z}}$ after $T$ iterations. Following this, DI2Iformer employs the IPR to complete I2I.

\section{Experiments}

\subsection{Experiment Settings}
 We apply our method to six I2I tasks separately: (a) inpainting, (b) real-world super-resolution (Real-World SR), (c) single-image super-resolution, (d) single-image motion deblurring, (e) semantic segmentation, and (f) depth estimation.  Regarding the diffusion models (DMs) settings in our DiffI2I for these tasks, we set the total timesteps $T$ to be 4, and the parameter $\beta_t$ in Eq.~\eqref{eq:mydiff2} (where $\alpha_t=1-\beta_t$) is linearly increased from $\beta_1=0.1$ to $\beta_T=0.99$ during the diffusion and reverse process. The task-specific settings will be explained in detail in the subsequent sections.

\subsection{Evaluation on Inpainting}
% We train and validate our DiffI2I$_{S2}$ on inpainting using the same settings of LaMa~\cite{LaMa}. Specifically, we adopt the Adam optimizer with learning rates 0.0002 and 0.0001 for DiffI2I and discriminator networks, respectively. All models are trained for 1M iterations with a batch size of 30. In addition, we use random crops of size 256$\times$256 to train DiffI2I on Places-Standard~\cite{places2} and CelebA-HQ~\cite{celeba} datasets. In testing, we use a fixed set of 2K validation and 30K testing samples from CelebA-HQ and
% Places-Standard. Moreover, we validate DiffI2I$_{S2}$ on crops of size 512$\times$512 and 256$\times$256 on Places and CelebA-HQ validation datasets, respectively.

We conduct training and validation of our DiffI2I for inpainting using the identical settings employed in the LaMa~\cite{LaMa}. Specifically, we opt for the Adam optimizer and set the learning rates to 0.0002 and 0.0001 for training the DiffI2I and the discriminator networks (used for GAN loss~\cite{WGAN-GP}), respectively. In the training process, all models are iterated over 1M times with a batch size of 30. Moreover, we train DiffI2I by employing random crops of size 256$\times$256 sourced from the Places-Standard~\cite{places2} (Places) and CelebA-HQ~\cite{celeba} datasets, respectively. During testing, we utilize fixed datasets comprising 2,000 validation samples and 30,000 testing samples from the CelebA-HQ and Places datasets separately. Furthermore, the validation of DiffI2I$_{S2}$ is conducted using crop sizes of 512$\times$512 for the Places dataset and 256$\times$256 for the CelebA-HQ dataset.

We evaluate our DiffI2I$_{S2}$ against SOTA inpainting methods, including ICT~\cite{ICT}, LaMa~\cite{LaMa}, RePaint~\cite{repaint}, using metrics such as LPIPS~\cite{LPIPS} and FID~\cite{FID} on validation datasets. Moreover, we compare DiffI2I$_{S2}$ with our baseline DiffIR~\cite{diffir}.  The quantitative results are presented in Tab.~\ref{tab:inpainting} and Fig.~\ref{fig:head}~(a). Our DiffI2I$_{S2}$ exhibits remarkable superiority over the other methods. 
In particular, DiffI2I$_{S2}$ outperforms the competitive LaMa by a substantial margin in terms of FID, achieving up to 0.3665 and 1.001 improvements with wide masks on the Places and CelebA-HQ datasets, respectively. Notably, this improvement is achieved while maintaining comparable levels of total parameters and runtime. Moreover, in comparison to the diffusion model (DM) based method RePaint~\cite{LDM}, DiffI2I$_{S2}$ demonstrates superior performance while utilizing significantly fewer resources—merely 4.5$\%$ of the parameters and 0.03$\%$ of the runtime. These results underscore the ability of DiffI2I to effectively leverage the data estimation capabilities of DM for impressive and efficient I2I outcomes. Furthermore, our DiffI2I surpasses our baseline model, DiffIR~\cite{diffir}. This substantial improvement validates the efficacy of our modified DI2Iformer architecture.

The qualitative results are shown in Fig.~\ref{fig:inpainting_show}. Notably, our DiffI2I$_{S2}$ demonstrates its capacity to generate remarkably realistic and coherent structures, along with intricate details, outperforming other competitive inpainting methods.

\subsection{Evaluation on Real-world Super-Resolution}

We train and validate our DiffI2I and baseline DiffIR~\cite{diffir} on real-world SR  using the same configuration as Real-ESRGAN~\cite{Real-ESRGAN}. This involves employing identical loss functions, as proposed by Real-ESRGAN~\cite{ESRGAN}, which incorporate perceptual and adversarial losses in addition to the fundamental $\mathcal{L}{1}$ loss. The learning rate for DiffI2I$_{S2}$ is set to $2\times10^{-4}$. During optimization, the Adam optimizer is employed with $\beta{1}=0.9$ and $\beta_{2}=0.99$. For both stages of training, we set the batch size to $64$, with the input patch size being $64$.

The evaluation of all methods is conducted on the dataset presented in the Real-World Super-Resolution challenge: NTIRE2020 Track1 and Track2~\cite{NTIRE2020} and AIM2019 Track2~\cite{Aim2019}. Moreover, we assess the performance of our DiffI2I model on the RealSRSet~\cite{RealSR}. As the NTIRE2020 Track1, AIM2019 Track2, and RealSRSet datasets offer paired validation sets, we employ LPIPS~\cite{LPIPS}, DISTS~\cite{DISTS}, and PSNR as evaluation metrics. We compare our DiffI2I and DiffIR~\cite{diffir} with competitive real-world SR methods, such as BSRGAN~\cite{BSRGAN}, Real-ESRGAN~\cite{Real-ESRGAN},  KDSR-GAN~\cite{KDSR}, and LDM~\cite{LDM}. 

The quantitative results are shown in Tab.~\ref{tab:real} and Fig.~\ref{fig:head}~(b). We can see that  DiffI2I$_{S2}$  outperforms SOTA real-world SR method KDSR$_{S}$-GAN on LPIPS, DISTS, and PSNR, consuming similar runtime. In addition, we can see that  DiffI2I$_{S2}$ outperforms classic real-world SR method Real-ESRGAN on LPIPS, DISTS, and PSNR, only consuming fewer runtime. Furthermore, compared with DM-based LDM~\cite{LDM}, DiffI2I$_{S2}$ achieve significant 1.84~dB PSNR and 0.182 LPIPS improvement on NTIRE2020 Track1 dataset consuming only $3.6\%$ runtime.

\begin{figure*}[t]
 \setlength{\fsdurthree}{0mm}
    \LARGE
    \centering
   \resizebox{1\linewidth}{!}{
        \begin{tabular}{cc}
            \begin{adjustbox}{valign=t}
                \Large
                \begin{tabular}{c}
                    \includegraphics[height=0.620\textwidth]{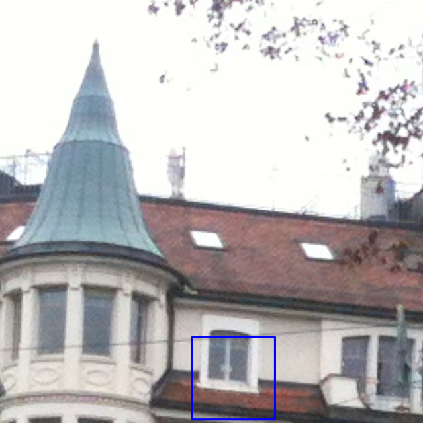} 
                \end{tabular}
                
            \end{adjustbox}
            
            \begin{adjustbox}{valign=t}
                \begin{tabular}{cccc}
                    \includegraphics[width=\widthscalefive \textwidth]{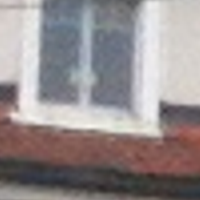} \hspace{\fsdurthree} &
                    \includegraphics[width=\widthscalefive \textwidth]{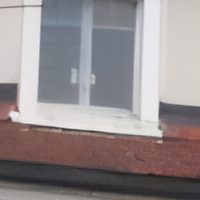} \hspace{\fsdurthree} &
                    \includegraphics[width=\widthscalefive \textwidth]{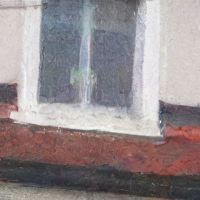} \hspace{\fsdurthree} 
                    \\
                    Input \hspace{\fsdurthree} &
                    \makecell{KDSR$_{S}$-GAN~\cite{KDSR}} \hspace{\fsdurthree} &
                    \makecell{LDM~\cite{LDM}} \hspace{\fsdurthree} 
                    \\
                    \includegraphics[width=\widthscalefive \textwidth]{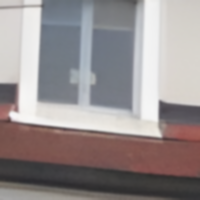} 
                    \hspace{\fsdurthree} &
                    \includegraphics[width=\widthscalefive \textwidth]{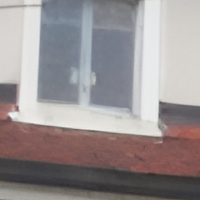} \hspace{\fsdurthree} &
                    \includegraphics[width=\widthscalefive \textwidth]{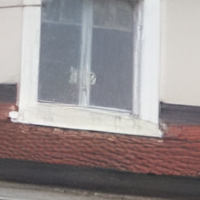} \hspace{\fsdurthree}  
                    \\ 
                    Real-ESRGAN~\cite{Real-ESRGAN} \hspace{\fsdurthree} &
                    \makecell{DiffIR$_{S2}$ (Ours)} \hspace{\fsdurthree} &
                    \makecell{DiffI2I$_{S2}$ (Ours)} \hspace{\fsdurthree} 
                \end{tabular}
            \end{adjustbox}
            
            \begin{adjustbox}{valign=t}
                \Large
                \begin{tabular}{c}
                    \includegraphics[height=0.620\textwidth]{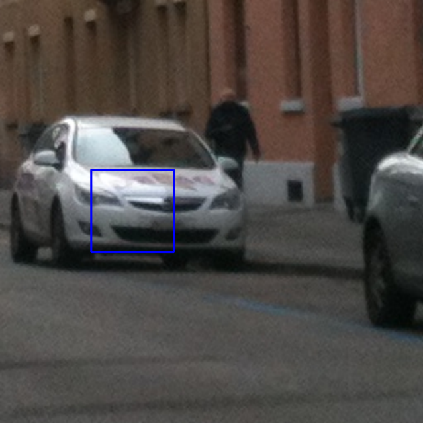} 
                \end{tabular}
                
            \end{adjustbox}
            
            \begin{adjustbox}{valign=t}
                \begin{tabular}{cccc}
                    \includegraphics[width=\widthscalefive \textwidth]{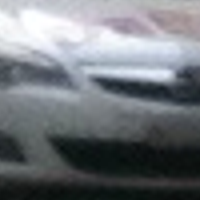} \hspace{\fsdurthree} &
                    \includegraphics[width=\widthscalefive \textwidth]{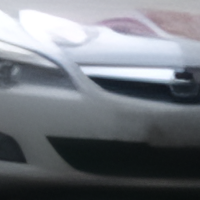} \hspace{\fsdurthree} &
                    \includegraphics[width=\widthscalefive \textwidth]{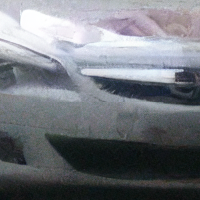} \hspace{\fsdurthree} 
                    \\
                    Input \hspace{\fsdurthree} &
                    \makecell{KDSR$_{S}$-GAN~\cite{KDSR}} \hspace{\fsdurthree} &
                    \makecell{LDM~\cite{LDM}} \hspace{\fsdurthree} 
                    \\
                    \includegraphics[width=\widthscalefive \textwidth]{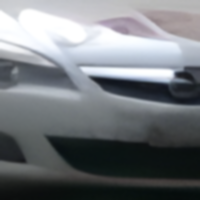} 
                    \hspace{\fsdurthree} &
                    \includegraphics[width=\widthscalefive \textwidth]{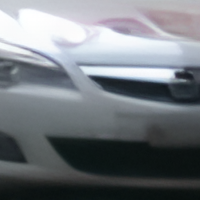} \hspace{\fsdurthree} &
                    \includegraphics[width=\widthscalefive \textwidth]{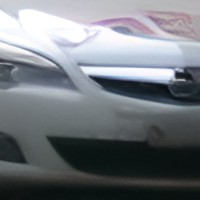} \hspace{\fsdurthree}  
                    \\ 
                    Real-ESRGAN~\cite{Real-ESRGAN} \hspace{\fsdurthree} &
                    \makecell{DiffIR$_{S2}$ (Ours)} \hspace{\fsdurthree} &
                    \makecell{DiffI2I$_{S2}$ (Ours)} \hspace{\fsdurthree} 
                \end{tabular}
            \end{adjustbox}
       \\
                   \begin{adjustbox}{valign=t}
                \Large
                \begin{tabular}{c}
                    \includegraphics[height=0.620\textwidth]{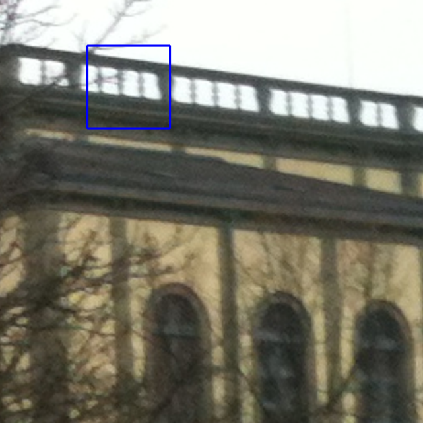} 
                \end{tabular}
                
            \end{adjustbox}
            
            \begin{adjustbox}{valign=t}
                \begin{tabular}{cccc}
                    \includegraphics[width=\widthscalefive \textwidth]{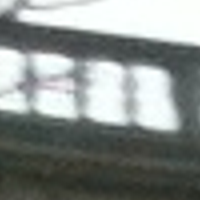} \hspace{\fsdurthree} &
                    \includegraphics[width=\widthscalefive \textwidth]{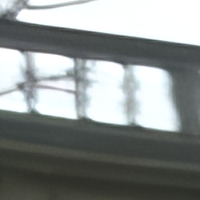} \hspace{\fsdurthree} &
                    \includegraphics[width=\widthscalefive \textwidth]{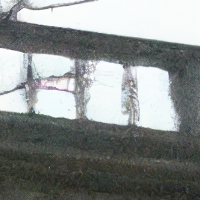} \hspace{\fsdurthree} 
                    \\
                    Input \hspace{\fsdurthree} &
                    \makecell{KDSR$_{S}$-GAN~\cite{KDSR}} \hspace{\fsdurthree} &
                    \makecell{LDM~\cite{LDM}} \hspace{\fsdurthree} 
                    \\
                    \includegraphics[width=\widthscalefive \textwidth]{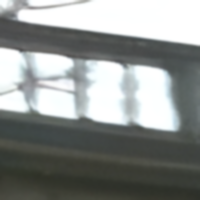} 
                    \hspace{\fsdurthree} &
                    \includegraphics[width=\widthscalefive \textwidth]{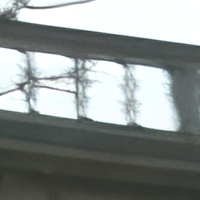} \hspace{\fsdurthree} &
                    \includegraphics[width=\widthscalefive \textwidth]{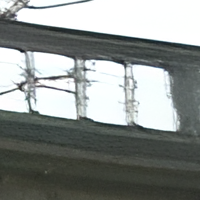} \hspace{\fsdurthree}  
                    \\ 
                    Real-ESRGAN~\cite{Real-ESRGAN} \hspace{\fsdurthree} &
                    \makecell{DiffIR$_{S2}$ (Ours)} \hspace{\fsdurthree} &
                    \makecell{DiffI2I$_{S2}$ (Ours)} \hspace{\fsdurthree} 
                \end{tabular}
            \end{adjustbox}
            
            \begin{adjustbox}{valign=t}
                \Large
                \begin{tabular}{c}
                    \includegraphics[height=0.620\textwidth]{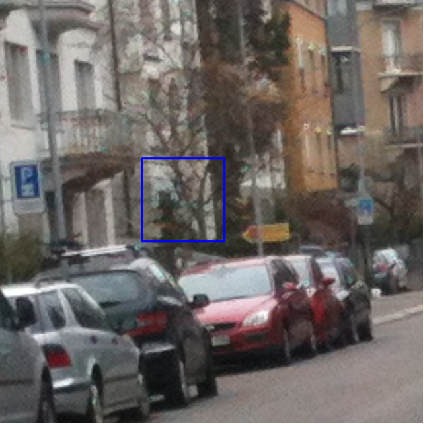} 
                \end{tabular}
                
            \end{adjustbox}
            
            \begin{adjustbox}{valign=t}
                \begin{tabular}{cccc}
                    \includegraphics[width=\widthscalefive \textwidth]{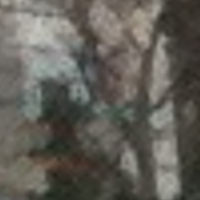} \hspace{\fsdurthree} &
                    \includegraphics[width=\widthscalefive \textwidth]{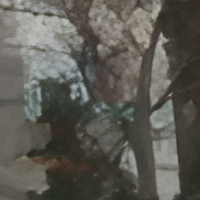} \hspace{\fsdurthree} &
                    \includegraphics[width=\widthscalefive \textwidth]{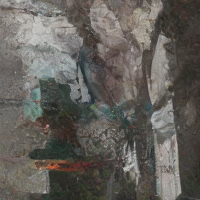} \hspace{\fsdurthree} 
                    \\
                    Input \hspace{\fsdurthree} &
                    \makecell{KDSR$_{S}$-GAN~\cite{KDSR}} \hspace{\fsdurthree} &
                    \makecell{LDM~\cite{LDM}} \hspace{\fsdurthree} 
                    \\
                    \includegraphics[width=\widthscalefive \textwidth]{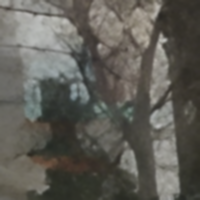} 
                    \hspace{\fsdurthree} &
                    \includegraphics[width=\widthscalefive \textwidth]{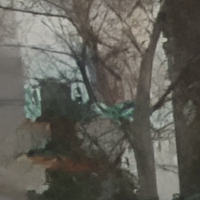} \hspace{\fsdurthree} &
                    \includegraphics[width=\widthscalefive \textwidth]{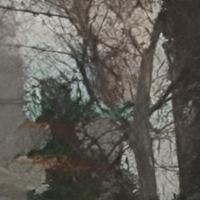} \hspace{\fsdurthree}  
                    \\ 
                    Real-ESRGAN~\cite{Real-ESRGAN} \hspace{\fsdurthree} &
                    \makecell{DiffIR$_{S2}$ (Ours)} \hspace{\fsdurthree} &
                    \makecell{DiffI2I$_{S2}$ (Ours)} \hspace{\fsdurthree} 
                \end{tabular}
            \end{adjustbox}

        \end{tabular}
    }
    \caption{ Visual comparison of 4$\times$ \textbf{real-world super-resolution} methods. Zoom-in for better details.}
    \vspace{-2mm}
    \label{fig:sup_realSR_show}
\end{figure*}

We have additionally provided visualizations of our results on NTIRE2020 Track2, a dataset captured using smartphones. These qualitative outcomes are illustrated in Fig.~\ref{fig:sup_realSR_show}. It is  apparent that DiffI2I$_{S2}$ excels in delivering the most impressive performance among all compared methods.

\begin{table*}[t]
  \centering
  \caption{Quantitative Comparison on 4$\times$ \textbf{real-world super-resolution} benchmarks.  The best and second-best performances are highlighted in bold and underlined, respectively. Notably, the lower trio of methods, shaded in gray, incorporate the diffusion model.}
\resizebox{1\linewidth}{!}{
    \begin{tabular}{l|ccc|ccc|ccc}
    \toprule[0.2em]
    \multirow{2}[2]{*}{Methods} & \multicolumn{3}{c|}{RealSRSet~\cite{RealSR}} & \multicolumn{3}{c|}{NTIRE2020 Track1~\cite{NTIRE2020}} & \multicolumn{3}{c}{AIM2019 Track2~\cite{Aim2019}} \\
          & LPIPS$\downarrow$ & DISTS$\downarrow$ & PSNR$\uparrow$  & LPIPS$\downarrow$ & DISTS$\downarrow$ & PSNR$\uparrow$  & LPIPS$\downarrow$ & DISTS$\downarrow$ & PSNR$\uparrow$ \\
    \midrule[0.2em]
    BSRGAN~\cite{BSRGAN} & 0.3648 & 0.1676 & 26.90  & 0.3691 & 0.1368 & 26.75 & 0.4048 & 0.1811 & 24.20 \\
    Real-ESRGAN~\cite{Real-ESRGAN} & 0.3629 & 0.1609 & 26.07 & 0.3471 & 0.1326 & 26.40  & 0.3956 & 0.1719 & 23.89 \\
    KDSR-GAN~\cite{KDSR} & 0.3610 & 0.1627 & 27.18 & 0.3198 & 0.1252 & 27.12 & 0.3758 & 0.1684 & 24.22 \\
    \rowcolor{lightgray}
    LDM~\cite{LDM}   & 0.4369 & 0.1982 & 26.37 & 0.4763 & 0.1844 & 25.68 & 0.5082 & 0.1972 & 22.63 \\
    \rowcolor{lightgray}
    DiffIR$_{S2}$~\cite{diffir} (Ours)& \underline{0.3527} & \underline{0.1588} & \underline{27.65} & \underline{0.3088} & \underline{0.1131} & \underline{27.31} & \underline{0.3650} & \underline{0.1718} & \underline{23.88} \\
    \rowcolor{lightgray}
    DiffI2I$_{S2}$ (Ours)&   \textbf{0.3457} & \textbf{0.1448} &	\textbf{27.77}	& \textbf{0.2943} & \textbf{0.1038} &	\textbf{27.52}	& \textbf{0.3476} &	\textbf{0.1533} &	\textbf{24.12} \\ 

    \bottomrule[0.2em]
    \end{tabular}%
    }
    \label{tab:real}%
\end{table*}%

\begin{table*}[t]
  \centering
  \caption{Quantitative evaluation ( LPIPS/DISTS). for \textbf{single image super-resolution} on benchmark datasets. The best and second-best performances are highlighted in bold and underlined, respectively. The bottom four methods marked in gray adopt the diffusion model.}
   \resizebox{1\linewidth}{!}{
    \begin{tabular}{l|cccccccccc}
    \toprule[0.2em]
    \multirow{2}[2]{*}{\textbf{Method}} & \multicolumn{2}{c}{\textbf{Set14}~\cite{Set14}} & \multicolumn{2}{c}{\textbf{Urban100}~\cite{Urban100}} & \multicolumn{2}{c}{\textbf{Manga109}~\cite{Manga109}} & \multicolumn{2}{c}{\textbf{General100}~\cite{general100}} & \multicolumn{2}{c}{\textbf{DIV2K100}~\cite{DIV2K}} \\
          & PSNR $\uparrow$& LPIPS $\downarrow$& PSNR $\uparrow$& LPIPS $\downarrow$& PSNR $\uparrow$& LPIPS $\downarrow$& PSNR $\uparrow$& LPIPS $\downarrow$& PSNR $\uparrow$& LPIPS $\downarrow$\\
    \midrule
    SFTGAN~\cite{SFTGAN} &    26.74  & 0.1313  & 24.34  & 0.1343  & 28.17  & 0.0716  & 29.16  & 0.0947  & 28.09  & 0.1331  \\
    SRGAN~\cite{SRGAN} &     26.84  & 0.1327  & 24.41  & 0.1439  & 28.11  & 0.0707  & 29.33  & 0.0964  & 28.17  & 0.1257  \\
    ESRGAN~\cite{ESRGAN} &    26.59  & 0.1241  & 24.37  & 0.1229  & 28.41  & 0.0649  & 29.43  & 0.0879  & 28.18  & 0.1154  \\
    USRGAN~\cite{USRGAN} &     27.41  & 0.1347  & 24.89  & 0.1330  & 28.75  & 0.0630  & 30.00  & 0.0937  & 28.79  & 0.1325  \\
    SPSR~\cite{SPSR}  & 26.86  & 0.1207  & 24.80  & 0.1184  & 28.56  & 0.0672  & 29.42  & 0.0862  & 28.18  & 0.1099  \\
    BebyGAN~\cite{BebyGAN} & 27.09  & 0.1157  & 25.23  & 0.1096  & 29.19  & 0.0529  & 29.95  & 0.0778  & 28.62  & 0.1022  \\
    \rowcolor{lightgray}
    LDM~\cite{LDM}   & 25.62  & 0.2034  & 23.36  & 0.1816  & 25.87  & 0.1321  & 27.17  & 0.1655  & 26.66  & 0.1939  \\
    \rowcolor{lightgray}
    SRDiff~\cite{srdiff}   & 27.14	& 0.1450 &	25.12 &	0.1379 &	28.67	& 0.0665 &	29.83 &	0.1009 &	28.58 &	0.1293 \\
    \rowcolor{lightgray}
    DiffIR$_{S2}$~\cite{diffir} (Ours) & \textbf{27.73}  & \underline{0.1117}  & \underline{26.05}  & \underline{0.1007}  & \underline{30.32}  & \underline{0.0463}  & \textbf{30.58}  & \underline{0.0717}  & \underline{29.13}  & \textbf{0.0871}  \\
    \rowcolor{lightgray}
     DiffI2I$_{S2}$ (Ours) &   	 \underline{27.46} &	\textbf{0.1094}		& \underline{25.99} &	\textbf{0.0925} &	\textbf{30.57} &	\textbf{0.0462}	&	\underline{30.39} &	\textbf{0.0666}	& \textbf{29.26} & \underline{0.0929} \\

    \bottomrule[0.2em]
    \end{tabular}%
    }
  \label{tab:SR}%

\end{table*}%

\begin{figure*}[t]

    % \newlength\fsdurthree
    \setlength{\fsdurthree}{0mm}
    \LARGE
    \centering
   \resizebox{1\linewidth}{!}{
        \begin{tabular}{cc}
            \begin{adjustbox}{valign=t}
                \Large
                \begin{tabular}{c}
                    \includegraphics[height=0.44\textwidth]{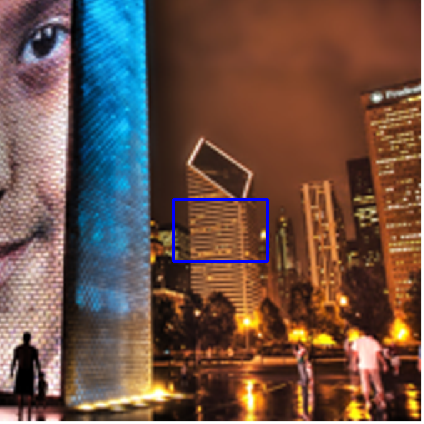} 
                \end{tabular}
                
            \end{adjustbox}
            
            \begin{adjustbox}{valign=t}
                \begin{tabular}{cccc}
                    \includegraphics[width=\widthscalefive \textwidth]{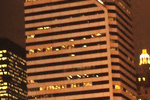} \hspace{\fsdurthree} &
                    \includegraphics[width=\widthscalefive \textwidth]{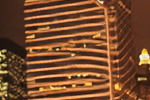} \hspace{\fsdurthree} &
                    \includegraphics[width=\widthscalefive \textwidth]{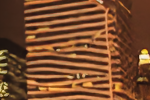} \hspace{\fsdurthree} 
                    \\
                    GT \hspace{\fsdurthree} &
                    \makecell{BebyGAN} \hspace{\fsdurthree} &
                    \makecell{SRDiff} \hspace{\fsdurthree} 
                    \\
                    \includegraphics[width=\widthscalefive \textwidth]{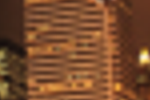} 
                    \hspace{\fsdurthree} &
                    \includegraphics[width=\widthscalefive \textwidth]{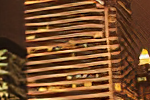} \hspace{\fsdurthree} &
                    \includegraphics[width=\widthscalefive \textwidth]{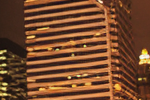} \hspace{\fsdurthree}  
                    \\ 
                    Input \hspace{\fsdurthree} &
                     \makecell{DiffIR$_{S2}$ (Ours)} \hspace{\fsdurthree} &
                    \makecell{DiffI2I$_{S2}$ (Ours)} \hspace{\fsdurthree} 
                \end{tabular}
            \end{adjustbox}

            \begin{adjustbox}{valign=t}
                \Large
                \begin{tabular}{c}
                    \includegraphics[height=0.44\textwidth]{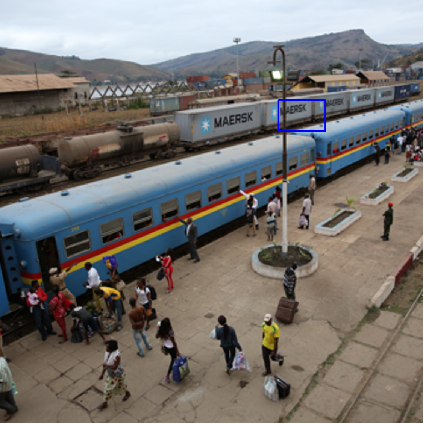} 
                \end{tabular}
                
            \end{adjustbox}
            
            \begin{adjustbox}{valign=t}
                \begin{tabular}{cccc}
                    \includegraphics[width=\widthscalefive \textwidth]{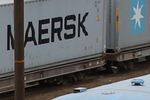} \hspace{\fsdurthree} &
                    \includegraphics[width=\widthscalefive \textwidth]{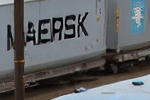} \hspace{\fsdurthree} &
                    \includegraphics[width=\widthscalefive \textwidth]{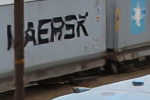} \hspace{\fsdurthree} 
                    \\
                    GT \hspace{\fsdurthree} &
                    \makecell{BebyGAN} \hspace{\fsdurthree} &
                    \makecell{SRDiff} \hspace{\fsdurthree} 
                    \\
                    \includegraphics[width=\widthscalefive \textwidth]{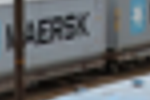} 
                    \hspace{\fsdurthree} &
                    \includegraphics[width=\widthscalefive \textwidth]{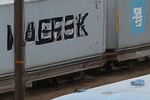} \hspace{\fsdurthree} &
                    \includegraphics[width=\widthscalefive \textwidth]{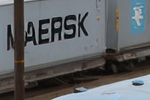} \hspace{\fsdurthree}  
                    \\ 
                    Input \hspace{\fsdurthree} &
                     \makecell{DiffIR$_{S2}$ (Ours)} \hspace{\fsdurthree} &
                    \makecell{DiffI2I$_{S2}$ (Ours)} \hspace{\fsdurthree} 
                \end{tabular}
            \end{adjustbox}
            \\
            \begin{adjustbox}{valign=t}
                \Large
                \begin{tabular}{c}
                    \includegraphics[height=0.44\textwidth]{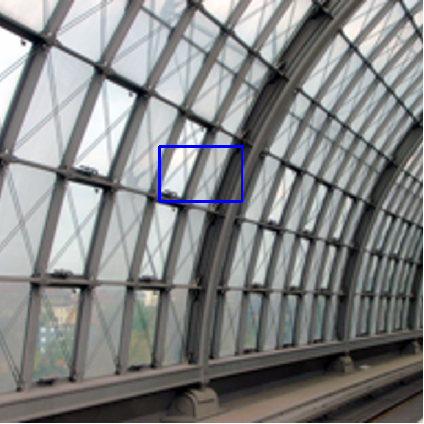} 
                \end{tabular}
                
            \end{adjustbox}
            
            \begin{adjustbox}{valign=t}
                \begin{tabular}{cccc}
                    \includegraphics[width=\widthscalefive \textwidth]{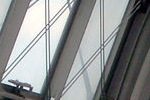} \hspace{\fsdurthree} &
                    \includegraphics[width=\widthscalefive \textwidth]{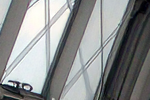} \hspace{\fsdurthree} &
                    \includegraphics[width=\widthscalefive \textwidth]{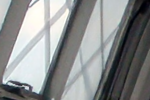} \hspace{\fsdurthree} 
                    \\
                    GT \hspace{\fsdurthree} &
                    \makecell{BebyGAN} \hspace{\fsdurthree} &
                    \makecell{SRDiff} \hspace{\fsdurthree} 
                    \\
                    \includegraphics[width=\widthscalefive \textwidth]{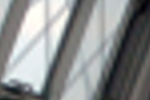} 
                    \hspace{\fsdurthree} &
                    \includegraphics[width=\widthscalefive \textwidth]{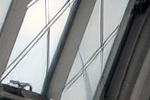} \hspace{\fsdurthree} &
                    \includegraphics[width=\widthscalefive \textwidth]{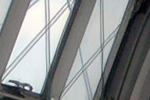} \hspace{\fsdurthree}  
                    \\ 
                    Input \hspace{\fsdurthree} &
                     \makecell{DiffIR$_{S2}$ (Ours)} \hspace{\fsdurthree} &
                    \makecell{DiffI2I$_{S2}$ (Ours)} \hspace{\fsdurthree} 
                \end{tabular}
            \end{adjustbox}

            \begin{adjustbox}{valign=t}
                \Large
                \begin{tabular}{c}
                    \includegraphics[height=0.44\textwidth]{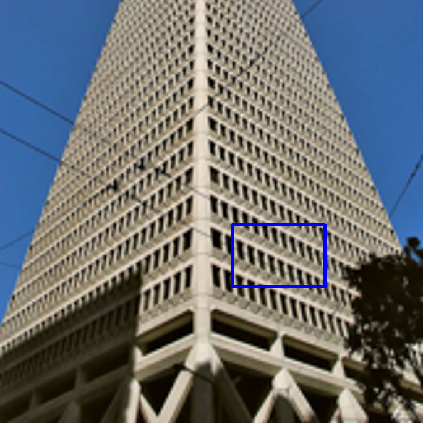} 
                \end{tabular}
                
            \end{adjustbox}
            
            \begin{adjustbox}{valign=t}
                \begin{tabular}{cccc}
                    \includegraphics[width=\widthscalefive \textwidth]{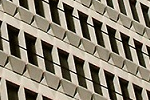} \hspace{\fsdurthree} &
                    \includegraphics[width=\widthscalefive \textwidth]{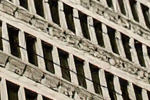} \hspace{\fsdurthree} &
                    \includegraphics[width=\widthscalefive \textwidth]{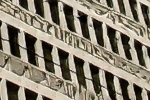} \hspace{\fsdurthree} 
                    \\
                    GT \hspace{\fsdurthree} &
                    \makecell{BebyGAN} \hspace{\fsdurthree} &
                    \makecell{SRDiff} \hspace{\fsdurthree} 
                    \\
                    \includegraphics[width=\widthscalefive \textwidth]{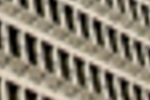} 
                    \hspace{\fsdurthree} &
                    \includegraphics[width=\widthscalefive \textwidth]{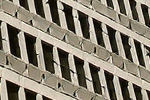} \hspace{\fsdurthree} &
                    \includegraphics[width=\widthscalefive \textwidth]{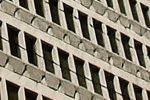} \hspace{\fsdurthree}  
                    \\ 
                    Input \hspace{\fsdurthree} &
                     \makecell{DiffIR$_{S2}$ (Ours)} \hspace{\fsdurthree} &
                    \makecell{DiffI2I$_{S2}$ (Ours)} \hspace{\fsdurthree} 
                \end{tabular}
            \end{adjustbox}
        \end{tabular}
    }
    \caption{ Visual comparison of 4$\times$ \textbf{image super-resolution} methods. Zoom-in for better details.}
    \label{fig:SR_show}
\end{figure*}

\subsection{Evaluation on Image Super-Resolution}

We train and validate our DiffI2I model for image super-resolution. Specifically, we train DiffI2I on DIV2K dataset~\cite{DIV2K} comprising 800 images, as well as the Flickr2K dataset~\cite{Flickr2K} consisting of 2650 images, all aimed at achieving a $4\times$ super-resolution. During training, we employ a batch size of 64, utilize input patches of size 64$\times$64, and set the number of iterations to 1M. To optimize our DiffI2I, we utilize the Adam optimizer with distinct learning rates: $0.0002$ for the DiffI2I network and $0.0001$ for the discriminator network (used in adversarial loss).

Subsequently, we subject our DiffI2I$_{S2}$ model, along with several other state-of-the-art  GAN-based SR methods, to evaluation across five benchmark datasets: Set5~\cite{Set5}, Set14~\cite{Set14}, General100~\cite{general100}, Urban100~\cite{Urban100}, and DIV2K100~\cite{DIV2K}. This evaluation employs two metrics, LPIPS~\cite{LPIPS} and PSNR, to gauge the performance of the models.
Tab.~\ref{tab:SR} and Fig.~\ref{fig:head}~(c) show the performance and runtime comparsion of  DiffI2I$_{S2}$ and DiffIR~\cite{diffir} with SOTA GAN-based SR methods: SFTGAN~\cite{SFTGAN}, SRGAN~\cite{SRGAN}, ESRGAN~\cite{ESRGAN}, USRGAN~\cite{USRGAN}, SPSR~\cite{SPSR}, and BebyGAN~\cite{BebyGAN}.  Compared with the competitive SR method BebyGAN, our DiffI2I$_{S2}$ surpasses it by LPIPS margin of up to 0.0112 and 0.0093 on General100 and DIV2K100 while merely consuming $82\%$ runtime. Moreover, it is notable that DiffI2I$_{S2}$ significantly outperforms the traditional DM-based method LDM while consuming $3.6\%$ runtime. 
Additionally, DiffI2I$_{S2}$ demonstrates a substantial performance improvement over SRDiff, all while utilizing only $1.3\%$ of the runtime. Moreover, comparing DiffIR$_{S2}$, DiffI2I$_{S2}$ achieves similar performance with fewer runtime. 

The qualitative results are displayed in Fig.~\ref{fig:SR_show}. Among them, DiffI2I$_{S2}$ stands out for its exceptional visual quality, encompassing a plethora of realistic details. These visual contrasts harmonize  with the corresponding quantitative findings, thus affirming the superiority of DiffI2I. Evidently, DiffI2I adequately harnesses the potency of DM, facilitating image super-resolution.

\begin{table}[t]
  \centering
  \caption{Quantitative evaluation for \textbf{single image motion deblurring} on benchmark datasets. The best and second-best performances are marked in bold and underlined, respectively. $^*$ represents using the TLC~\cite{TLC}.}
  \resizebox{1\linewidth}{!}{
    \begin{tabular}{l|cc|cc}
    \toprule[0.2em]
    \multirow{2}[2]{*}{\textbf{Method}} & \multicolumn{2}{c|}{\textbf{GoPro}~\cite{gopro}} & \multicolumn{2}{c}{\textbf{HIDE}~\cite{hide}} \\
          & \multicolumn{1}{c}{PSNR $\uparrow$} & \multicolumn{1}{c|}{SSIM $\uparrow$} & \multicolumn{1}{c}{PSNR $\uparrow$} & \multicolumn{1}{c}{SSIM $\uparrow$} \\
    \midrule[0.2em]
    Xu \etal~\cite{deblur-xu} & \multicolumn{1}{c}{21.00} & \multicolumn{1}{c|}{0.741} & \multicolumn{1}{c}{-} & \multicolumn{1}{c}{-} \\
    DeblurGAN~\cite{deblurgan} & \multicolumn{1}{c}{28.70} & \multicolumn{1}{c|}{0.858} & \multicolumn{1}{c}{24.51} & \multicolumn{1}{c}{0.871} \\
    Nah \etal~\cite{gopro} & \multicolumn{1}{c}{29.08} & \multicolumn{1}{c|}{0.914} & \multicolumn{1}{c}{25.73} & \multicolumn{1}{c}{0.874} \\
    Zhang \etal~\cite{deblur-zhang} & \multicolumn{1}{c}{29.19} & \multicolumn{1}{c|}{0.931} & \multicolumn{1}{c}{-} & \multicolumn{1}{c}{-} \\
    DeblurGAN-v2~\cite{deblurganv2} & \multicolumn{1}{c}{29.55} & \multicolumn{1}{c|}{0.934} & \multicolumn{1}{c}{26.61} & \multicolumn{1}{c}{0.875} \\
    SRN~\cite{SRN}   & \multicolumn{1}{c}{30.26} & \multicolumn{1}{c|}{0.934} & \multicolumn{1}{c}{28.36} & \multicolumn{1}{c}{0.915} \\
    Gao \etal~\cite{deblur-gao} & \multicolumn{1}{c}{30.90} & \multicolumn{1}{c|}{0.935} & \multicolumn{1}{c}{29.11} & \multicolumn{1}{c}{0.913} \\
    DBGAN~\cite{DBGAN} & \multicolumn{1}{c}{31.10} & \multicolumn{1}{c|}{0.942} & \multicolumn{1}{c}{28.94} & \multicolumn{1}{c}{0.915} \\
    MT-RNN~\cite{MT-RNN} & \multicolumn{1}{c}{31.15} & \multicolumn{1}{c|}{0.945} & \multicolumn{1}{c}{29.15} & \multicolumn{1}{c}{0.918} \\
    DMPHN~\cite{DMPHN} & \multicolumn{1}{c}{31.20} & \multicolumn{1}{c|}{0.940} & \multicolumn{1}{c}{29.09} & \multicolumn{1}{c}{0.924} \\
    Suin \etal~\cite{deblur-suin} & \multicolumn{1}{c}{31.85} & \multicolumn{1}{c|}{0.948} & \multicolumn{1}{c}{29.98} & \multicolumn{1}{c}{0.930} \\
    MIMO-Unet+~\cite{MIMO-Unet} & \multicolumn{1}{c}{32.45} & \multicolumn{1}{c|}{0.957} & \multicolumn{1}{c}{29.99} & \multicolumn{1}{c}{0.930} \\
    IPT~\cite{IPT}   & \multicolumn{1}{c}{32.52} & \multicolumn{1}{c|}{-} & \multicolumn{1}{c}{-} & \multicolumn{1}{c}{-} \\
    MPRNet~\cite{MPRNet} & \multicolumn{1}{c}{32.66} & \multicolumn{1}{c|}{0.959} & \multicolumn{1}{c}{30.96} & \multicolumn{1}{c}{0.939} \\
    Restormer~\cite{restormer} & \multicolumn{1}{c}{32.92} & \multicolumn{1}{c|}{0.961} & \multicolumn{1}{c}{31.22} & \multicolumn{1}{c}{0.942} \\
    NAFNet~\cite{NAFNet} & 33.08 & 0.965 &31.22& \underline{0.946}  \\
    NAFNet$^{*}$~\cite{NAFNet} & \underline{33.71} & \underline{0.967}&31.32&0.941  \\
    \midrule[0.2em]
    DiffIR$_{S2}$~\cite{diffir} (Ours) &   \multicolumn{1}{c}{33.20}    &       \multicolumn{1}{c|}{0.963}    & \multicolumn{1}{c}{\underline{31.55}} & \multicolumn{1}{c}{\textbf{0.947}} \\
    DiffI2I$_{S2}$ (Ours) & 33.25 & 0.964&\textbf{31.56}&\textbf{0.947}  \\
    DiffI2I$_{S2}^{*}$ (Ours) & \textbf{33.85} & \textbf{0.968}&31.42&0.943  \\
    \bottomrule[0.2em]
    \end{tabular}%
    }
  \label{tab:deblur}%
  \vspace{-2mm}
\end{table}%

\subsection{Evaluation on Image Motion Deblurring}

Following previous works in single-image motion deblurring~\cite{MIMO-Unet, MPRNet, restormer}, we train our DiffI2I utilizing only the $\mathcal{L}_{1}$ loss function to ensure fair comparisons. We train DiffI2I for 300,000 iterations, commencing with an initial learning rate of $2 \times 10^{-4}$, which gradually decreases to $1 \times 10^{-6}$ through the application of the cosine annealing~\cite{cosine}. In harmony with previous research~\cite{restormer}, we progressively the batch size during training. Specifically, our training process initiates with a patch size of $128 \times 128$ and a batch size of $64$. The patch size and batch size combination dynamically evolves to $[(160 \times 160, 40), (192 \times 192, 32), (256 \times 256, 16), (320 \times 320, 8), (384 \times 384, 8)]$ at the corresponding iterations of $[92K, 156K, 204K, 240K, 276K]$. We train DiffI2I on GoPro~\cite{gopro} dataset for image motion deblurring and evaluate DiffI2I  on two classic benchmarks (GoPro, HIDE~\cite{hide}).

We present a comprehensive comparison of DiffI2I$_{S2}$ with state-of-the-art image motion deblurring methods, including Restormer~\cite{restormer}, MPRNet~\cite{MPRNet}, IPT~\cite{IPT}, NAFNet~\cite{NAFNet}, and our baseline, DiffIR~\cite{diffir}.  The quantitative evaluation, measured in terms of PSNR and SSIM, is detailed in Tab.~\ref{tab:deblur}, while the runtime comparisons are illustrated in Fig.~\ref{fig:head}~(d). The results  underscore the superiority of DiffI2I$_{S2}$ over alternative motion deblurring methods.  Specifically, DiffI2I$_{S2}$  exhibits a remarkable performance advantage, surpassing IPT and MIMI-Unet+ by a substantial margin of 0.73dB and 0.8dB on the GoPro dataset, respectively. Furthermore, on both the GoPro and HIDE datasets, DiffI2I$_{S2}$ demonstrates a clear advantage over Restormer, achieving improvements of 0.33dB and 0.34dB, respectively, while only consuming  63$\%$ runtime.  Additionally, regardless of whether utilizing TLC~\cite{TLC}, our DiffI2I$_{S2}$ consistently outperforms NAFNet. Moreover, when compared to our own DiffIR$_{S2}$, DiffI2I$_{S2}$ proves its enhanced efficacy by delivering superior results with a mere 68$\%$ of the runtime. This demonstrates the effectiveness of our DiffI2I.

\subsection{Evaluation on Semantic Segmentation}
Following the pretraining details of seaformer~\cite{seaformer}, we adopt the network pre-trained on ImageNet-1K~\cite{imagenet} as the backbone. For semantic segmentation, the standard BatchNorm~\cite{batchnorm} layer is replaced by synchronized BatchNorm. Our model undergoes comprehensive training and evaluation on the ADE20K~\cite{ADE20K} dataset, which encompasses 20K/2K/3K images allocated for training, validation, and testing, encompassing 150 categories. Our method  aligns with the training settings elucidated in segformer~\cite{segformer}. Specifically, we adopt a 160K scheduler and maintain a batch size of 16 throughout the training process.  We employ the AdamW optimizer and regulate the learning rate at 2.5$\times10^{-4}$ while using a "poly" learning rate schedule with a factor of 1.0. To better accommodate the semantic segmentation, we use axial attention~\cite{seaformer} to replace the channel attention in the DA block of DI2Iformer to obatin the encoder.

We present a comparison of our DiffI2I$_{S2}$ against previous methods on the ADE20K validation set, as shown in Tab.~\ref{tab:seg}. Additionally, we offer a runtime comparison with state-of-the-art  semantic segmentation methods, illustrated in Fig.~\ref{fig:head} (e). Notably, our DiffI2I$_{S2}$ outperforms the other methods in terms of mIOU while demonstrating comparable or lower runtime consumption.

\begin{table}[t]
  \centering
  \caption{Quantitative evaluation for \textbf{semantic segmentation} on ADE20K~\cite{ADE20K} val set. The best and second-best performances are marked in bold and underlined, respectively.}
  \resizebox{1\linewidth}{!}{
    \begin{tabular}{llll}
    \toprule[0.2em]
    \textbf{Backbone} & \textbf{Decoder} & \textbf{Params (M)} & \textbf{mIOU$\uparrow$} \\
    \midrule[0.2em]
    ConvMLP-S~\cite{convmlp} & Semantic FPN~\cite{semanticFPN} & 12.8  & 35.8 \\
    EfficientNet~\cite{efficientnet} & DeepLabV3+~\cite{lite-aspp} & 17.1  & 36.2 \\
    MobileNetV2~\cite{mobilenetv2} & Lite-ASPP~\cite{lite-aspp} & 2.9   & 36.6 \\
    MiT-B0~\cite{segformer} & SegFormer & 3.8   & 37.4 \\
    TopFormer-S~\cite{segformer} & Simple Head & 3.1   & 36.5 \\
    SeaFormer-S~\cite{seaformer} & Light Head & 4.0     & \underline{38.1} \\
    DiffI2I$_{S2}$-S & Light Head & 4.4   & \textbf{38.8} \\
    \midrule
    ResNet18~\cite{resnet} & Lite-ASPP~\cite{lite-aspp} & 12.5  & 37.5 \\
    ShuffleNetV2-1.5x~\cite{shufflenet} & DeepLabV3+~\cite{lite-aspp} & 16.9  & 37.6 \\
    MobileNetV2~\cite{mobilenetv2} & DeepLabV3+~\cite{lite-aspp} & 15.4  & 38.1 \\
    MiT-B1~\cite{segformer} & SegFormer & 13.7  & 41.6 \\
    Seaformer-L~\cite{seaformer} & Light Head & 14.0    & \underline{42.7} \\
    DiffI2I$_{S2}$-L & Light Head & 12.9  & \textbf{43.5} \\
    \bottomrule[0.2em]
    \end{tabular}%
    }
  \label{tab:seg}%
  \vspace{-2mm}
\end{table}%

\subsection{Evaluation on Depth Estimation}

We train the DiffI2I using a batch size of 16 and a learning rate of $1\times10^{-4}$ for a total of $38.4K$ iterations. We employ the AdamW optimizer with  $(\beta_1, \beta_2) = (0.9, 0.999)$. Moreover, we implement a linear learning rate warm-up strategy during the initial 30$\%$ of iterations. Additionally, we incorporate a cosine annealing learning rate decay strategy to further refine the learning process. The effectiveness of our approach is validated through training and evaluation on the NYU-Depth-v2 dataset~\cite{YOLO}. To better accommodate the depth estimation and follow binsformer does, we use a pre-trained Swin transformer as the encoder of DI2Iformer.

We assess the performance of our DiffI2I$_{S2}$ by conducting a comparison with existing methods on the NYU-Depth-v2 validation dataset in Tab.~\ref{tab:depth}. Additionally, we present a runtime comparison with SOTA depth estimation methods, as depicted in Fig.~\ref{fig:head} (f). It is evident from our findings that DiffI2I$_{S2}$ not only outperforms the other methods in terms of performance but also exhibits lower or similar runtime, further establishing its superiority in depth estimation.

\begin{table}[t]
  \centering
  \caption{Quantitative evaluation for \textbf{depth estimation} on NYU-Depth-v2~\cite{YOLO}. The best and second-best performances are marked in bold and underlined, respectively. The last two methods, shaded in gray, utilize the diffusion model.}
  \resizebox{1\linewidth}{!}{
    \begin{tabular}{l|ccc|ccc}
    \toprule[0.2em]
    \textbf{Method} & \textbf{Rel.$\downarrow$}   & \textbf{RMSE$\downarrow$}  & \textbf{log$_{10}\downarrow$} & $\delta^1\uparrow$    & $\delta^2\uparrow$    & $\delta^3\uparrow$ \\
    \midrule[0.2em]
    VNL~\cite{VNL}   & 0.108  & 0.416  & 0.048  & 0.875  & 0.976  & 0.994  \\
    BTS~\cite{BTS}   & 0.113  & 0.407  & 0.049  & 0.871  & 0.977  & 0.995  \\
    DAV~\cite{DAV}   & 0.108  & 0.412  & -     & 0.882  & 0.980  & 0.996  \\
    PWA~\cite{PWA}   & 0.105  & 0.374  & 0.045  & 0.892  & 0.985  & \underline{0.997}  \\
    TransDepth~\cite{transdepth} & 0.106  & 0.365  & 0.045  & 0.900  & 0.983  & 0.996  \\
    Adabins~\cite{adabins} & 0.103  & 0.364  & 0.044  & 0.903  & 0.984  & \underline{0.997}  \\
    P3Depth~\cite{p3depth} & 0.104  & 0.356  & 0.043  & 0.898  & 0.981  & 0.996  \\
    DepthFormer~\cite{depthformer} & 0.096  & 0.339  & 0.041  & 0.921  & 0.989  & \textbf{0.998}  \\
    NeWCRFs~\cite{p3depth} & 0.095  & 0.334  & 0.041  & 0.922  & \textbf{0.992}  & \textbf{0.998}  \\
    % PixelFormer & 0.090  & 0.322  & 0.039  & 0.929  & 0.991  & 0.998  \\
    BinsFormer~\cite{binsformer} & \underline{0.094}  & \underline{0.330}  & \underline{0.040}  & \underline{0.925}  & 0.989  & \underline{0.997}  \\
    \rowcolor{lightgray}
    DDP~\cite{ddp}   & 0.101  & 0.336  & 0.042  & 0.923  & 0.990  & \textbf{0.998}  \\
    \rowcolor{lightgray}
    DiffI2I$_{S2}$ & \textbf{0.091} 	& \textbf{0.315} &	\textbf{0.038} & 	\textbf{0.933} & 	\underline{0.991} & 	\textbf{0.998}  \\
    \bottomrule[0.2em]
    \end{tabular}%
    }
  \label{tab:depth}%
\end{table}%

\section{Ablation Study}
\label{sec:ablation}
\noindent\textbf{Efficient diffusion model for I2I.}
 In this section, we undertake a comprehensive validation of the efficacy of various components within DiffI2I. These components include the DM, distinct training schemes for DM, as well as the impact of introducing variance noise into the DM (refer to Tab.~\ref{tab:DiffI2I}).

\textbf{(1)} DiffI2I$_{S2}$-V3 is actually the DiffI2I$_{S2}$ model used in Tab.~\ref{tab:inpainting}, and DiffI2I$_{S1}$ represents the first stage pretraining network that utilizes ground-truth images as inputs. A comparative analysis between DiffI2I$_{S1}$ and DiffI2I$_{S2}$-V3 reveals a noteworthy similarity in their LPIPS scores. This observation underscores the powerful data modeling capabilities of the DM in accurately predicting IPR.

\textbf{(2)} To provide additional evidence of the DM's efficacy, we intentionally omit its usage in DiffI2I$_{S2}$-V3, obtaining DiffI2I$_{S2}$-V1. A comparison between DiffI2I$_{S2}$-V1 and DiffI2I$_{S2}$-V3 reveals a marked performance disparity, with DiffI2I$_{S2}$-V3 (leveraging DM) notably surpassing DiffI2I$_{S2}$-V1. This outcome underscores the pivotal role of the IPR learned by the DM in effectively guiding DI2Iformer for I2I.

\textbf{(3)} To investigate optimal training schemes for DM, we contrast two distinct approaches: traditional DM optimization and our proposed joint optimization. The traditional DM optimization, as utilized in previous works such as~\cite{LDM,DDPM}, necessitates numerous iterations for estimating large images or feature maps. Consequently, it resorts to random timestep sampling for optimizing the denoising network, rendering it unable to synchronize optimization with subsequent components, specifically the decoder (i.e., the DI2Iformer in our study). 
In contrast, DiffI2I relies on DM solely to derive a compact one-dimensional vector IPR. This permits fewer iterations and obtains notably accurate outcomes. Hence, we adopt a joint optimization scheme that runs all iterations of the denoising network to obtain IPR, facilitating mutual optimization with DI2Iformer. A comparative analysis between DiffI2I$_{S2}$-V2 and DiffI2I$_{S2}$-V3 reveals a significant performance disparity, with DiffI2I$_{S2}$-V3 outperforming DiffI2I$_{S2}$-V2. This significant improvement underscores the efficacy of our proposed joint optimization for training DM. This joint optimization is especially valuable as even minor inaccuracies in DM's IPR estimation can detrimentally affect DI2Iformer's performance. By training DM and DI2Iformer jointly, we mitigate this challenge.

\textbf{(4)} 
In traditional DM methods, a common practice involves introducing variance noise during the reverse DM process (as shown in Eq.~\eqref{eq:diff3}) to enhance the realism of generated images.
Diverging from these traditional DMs, our DiffI2I takes a departure by eliminating the incorporation of noise into the DM process. We undertake experiments to validate the efficacy of this approach. Specifically, within the DiffI2I$_{S2}$-V4, we introduce noise during the reverse DM process. As shown in Tab.~\ref{tab:DiffI2I}, our noise-free DiffI2I$_{S2}$-V3 outperforms the noise-inclusive DiffI2I$_{S2}$-V4 model in terms of performance. This observation suggests that abstaining from the addition of noise leads to heightened accuracy in estimating IPR.

\begin{table*}[htbp]
  \centering
  \caption{
FID results evaluated on the CelebA-HQ dataset for inpainting. The evaluation encompassed both performance  and runtime, all conducted using inputs with the resolution of 256x256}
  \resizebox{1\linewidth}{!}{
    \begin{tabular}{l|c|ccccc|c}
    \toprule[0.2em]
    \multirow{2}[4]{*}{\textbf{Method}} & \multirow{2}[4]{*}{\textbf{\shortstack{Runtime (ms)}}} & \multirow{2}[4]{*}{\textbf{GT}} & \multirow{2}[4]{*}{\textbf{DM}} & \multicolumn{2}{c}{\textbf{Training Schemes}} & \multirow{2}[4]{*}{\textbf{\shortstack{Inserting\\ Noise}} } & \multirow{2}[4]{*}{\textbf{CelebA-HQ (FID$\downarrow$)} } \\
\cmidrule{5-6}          &       &       &       & \textbf{\shortstack{Traditional DM \\Optimization}}  & \textbf{\shortstack{Joint \\Optimization}} &       &  \\
    \midrule
    DiffI2I$_{S1}$ &   28.1    & \Checkmark     & \XSolidBrush     & \XSolidBrush     & \XSolidBrush     & \XSolidBrush     & 4.4826 \\
    \midrule
    DiffI2I$_{S2}$-V1 &   28.6    & \XSolidBrush     & \XSolidBrush     & \XSolidBrush     & \XSolidBrush     & \XSolidBrush     &  5.2452\\
    DiffI2I$_{S2}$-V2 &    28.6   & \XSolidBrush     & \Checkmark     & \Checkmark     & \XSolidBrush     & \XSolidBrush     &  5.5221\\
    DiffI2I$_{S2}$-V3 (Ours) &    28.6   & \XSolidBrush     & \Checkmark     & \XSolidBrush     & \Checkmark     & \XSolidBrush     &  4.7013\\
    DiffI2I$_{S2}$-V4 &   28.6    & \XSolidBrush     & \Checkmark     & \XSolidBrush     & \Checkmark     & \Checkmark     & 4.7796 \\
    \bottomrule[0.2em]
    \end{tabular}%
}
  \label{tab:DiffI2I}%
\end{table*}%

\begin{table}[t]
  \centering
  \caption{ Various DM loss functions comparison (FID) in inpainting.}
  \resizebox{1\linewidth}{!}{
    \begin{tabular}{l|ccc}
    \toprule[0.2em]
    Loss functions  & $\mathcal{L}_{diff}$ (Eq.~\eqref{eq:diff})   & $\mathcal{L}_{2}$ (Eq.~\eqref{eq:l2})    & $\mathcal{L}_{kl}$ (Eq.~\eqref{eq:kl}) \\
    \midrule
    CelebA-HQ (FID$\downarrow$) & 4.7013     & 4.7518     & 4.7963 \\
    \bottomrule[0.2em]
    \end{tabular}%
    }
  \label{tab:loss}%
\end{table}%

\begin{figure}[t]
	\centering
	\includegraphics[height=4.5cm]{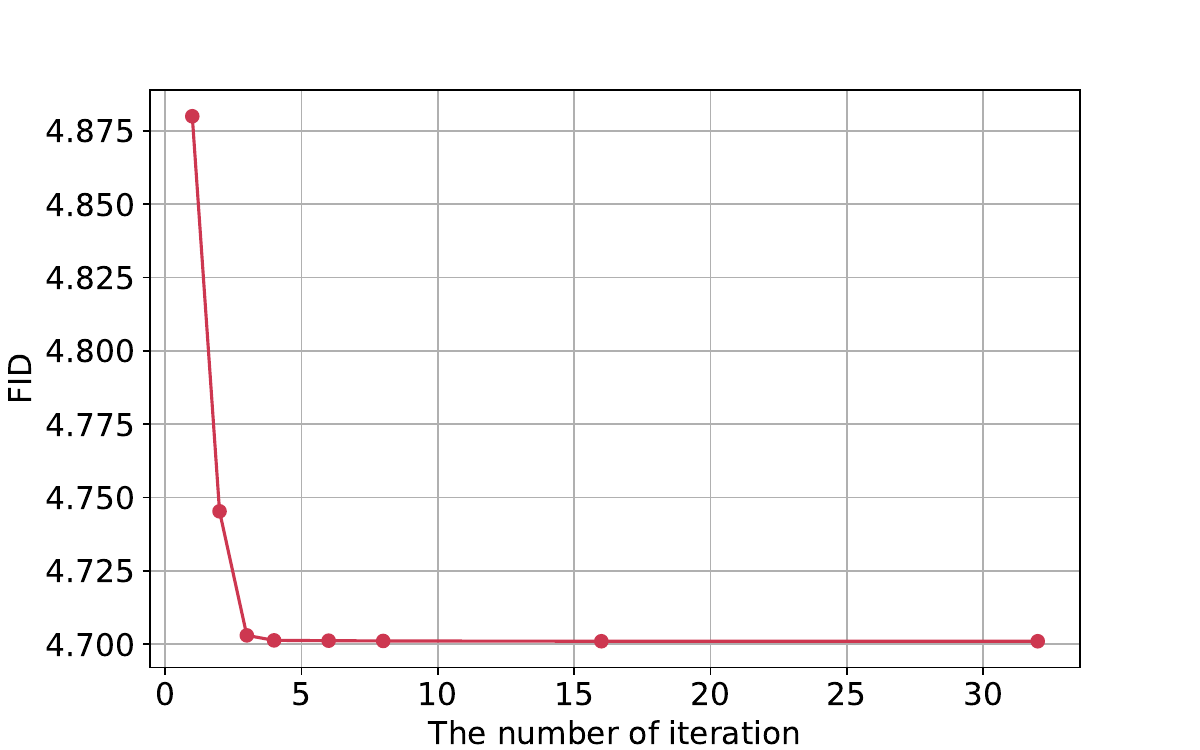}
	\caption{Ablation study of the number of iterations in DM.  }
	\label{fig:iter}
\end{figure}

\noindent\textbf{The loss functions for DM.}
We investigate the optimal loss function for guiding both the denoising network and CPEN$_{S2}$ in accurately estimating the IPR from input images. In this context, we introduce three distinct loss functions:
\textbf{(1)}
We define $\mathcal{L}_{diff}$ and outlined in Eq.~\eqref{eq:diff} to serve as our optimization criterion.
\textbf{(2)}
To quantify the extent of estimation error, we employ the $\mathcal{L}_{2}$ loss (Eq.~\eqref{eq:l2}).
\textbf{(3)}
We utilizes the Kullback-Leibler (KL) divergence to gauge the similarity between distributions, denoted as $\mathcal{L}_{kl}$ (Eq.~\eqref{eq:kl}).
\begin{equation}
\label{eq:l2}
\mathcal{L}_{2}=\frac{1}{4C^{\prime}}\sum_{i=1}^{4C^{\prime}}\left(\hat{\mathbf{Z}}(i)-\mathbf{Z}(i)\right)^{2}, 
\end{equation}
\begin{equation}
\label{eq:kl}
\begin{aligned}
\mathcal{L}_{kl} &= \sum_{i=1}^{4C^{\prime}} \mathbf{Z}_{norm}(i) \log \left(\frac{\mathbf{Z}_{norm}(i)}{\hat{\mathbf{Z}}_{norm}(i)}\right),
\end{aligned}
\end{equation}
where $\mathbf{Z}$ and $\hat{\mathbf{Z}}\in \mathbb{R}^{4C^{\prime}}$  represent IPRs extracted by DiffI2I$_{S1}$ and DiffI2I$_{S2}$, respectively.  $\hat{\mathbf{Z}}_{norm}$ and $\mathbf{Z}_{norm} \in \mathbb{R}^{4C^{\prime}}$ denote the results of applying the softmax operation to normalize $\hat{\mathbf{Z}}$ and $\mathbf{Z}$ independently. We employ these three loss functions on DiffI2I$_{S2}$ individually, aiming to directly predict accurate IPR values from input images. Subsequently, we assess their performance in the inpainting task using the CelebA-HQ dataset. The outcome of these evaluations is presented in Tab.~\ref{tab:loss}. Notably, our results demonstrate that $\mathcal{L}_{diff}$ outperforms both $\mathcal{L}_{2}$ and $\mathcal{L}_{kl}$.

\begin{table}[t]
  \centering
  \caption{Comparison on 4$\times$ real-world SR efficient DM methods}
  \resizebox{1.\linewidth}{!}{
    \begin{tabular}{c|c|ccc}
    \toprule[0.2em]
    Method & Runtime (ms) & LPIPS$\downarrow$ & DISTS$\downarrow$ & PSNR$\uparrow$ \\
    \midrule
    LDM~\cite{LDM} & 4706.18 & 0.4763 & 0.1844 & 25.68 \\
    LDM+DPM-Sovler~\cite{dpm-solver} & 3411.59 & 0.5003 & 0.1871 & 24.31 \\
    DiffI2I$_{S2}$ (Ours) & 170.4 & 0.2943 & 0.1038 & 27.52 \\
    \bottomrule[0.2em]
    \end{tabular}%
    }
   \label{tab:efficient}%
\end{table}%

\noindent\textbf{Impact of the number of iterations.}
In this part, we delve into the impact of the number of iterations in the DiffI2I$_{S2}$ on its performance. We employ varying iteration counts within DiffI2I$_{S2}$ and fine-tune the parameter $\beta_{t}$ ($\alpha_t=1-\beta_{t}$) as depicted in Eq.~\eqref{eq:mydiff1}, ensuring that $\mathbf{Z}$ evolves into Gaussian noise $\mathbf{Z}_{T} \sim \mathcal{N}(0,1)$ through the diffusion process (i.e., as $\bar{\alpha}_{T} \rightarrow 0$).
The results are illustrated in Fig.~\ref{fig:iter}. Remarkably, as the number of iterations reaches 3, the performance of DiffI2I$_{S2}$ experiences a substantial enhancement. Beyond 4 iterations, DiffI2I$_{S2}$ stabilizes, effectively reaching its upper-performance limit. Furthermore, it is evident that our DiffI2I$_{S2}$ exhibits a swifter convergence rate compared to traditional DMs, which typically necessitate over 200 iterations. This advantageous acceleration arises from our focused application of DM on the IPR, a compressed one-dimensional vector.

\noindent\textbf{Comparison with efficient DM methods.} 
We conduct a comparative analysis involving LDM~\cite{LDM}, which is an efficient DM method and is accelerated by DDIM~\cite{DDIM}. Additionally, we leverage the  DPM-Solver~\cite{dpm-solver} to optimize the efficiency of LDM. It is officially adopted for accelerating both LDM and Stable Diffusion.
As illustrated in Tab.~\ref{tab:efficient}, our proposed DiffI2I framework, tailored for I2I, demonstrates remarkable superiority over DPM-Solver, showcasing a performance enhancement while being 20 times more efficient.

\noindent\textbf{Whether concatenate input and GT images.} Our experimental evaluations are conducted on the NTIRE2020-track1 dataset, which is a real-world SR dataset.
As indicated in Tab.~\ref{tab:GT}, two distinct inputs are considered for the CPEN$_{S1}$: solely the ground truth (GT) images, and the concatenation of both GT and input images. It is evident that employing the concatenation of GT and input images as input to CPEN$_{S1}$ yields superior performance. This approach facilitates the DM in comprehending the disparity between the input and GT images, thereby enhancing its ability to provide more effective guidance for I2I.

\noindent\textbf{The effect of resolutions.} 
Our experimental evaluations are conducted on the Single Image Super-Resolution (SISR) dataset, precisely the DIV2K100 dataset, which is resized to a variety of resolutions. The results, displayed in Tab.~\ref{tab:resolution}, unequivocally showcase DiffI2I$_{S2}$'s superior performance across different resolutions, surpassing the ByteGAN and SRDiff. These results underscore the exceptional adaptability of our DiffI2I$_{S2}$ to diverse resolution settings.

\begin{table}[t]
  \centering
  \caption{Whether concatenate GT and input images?}
    \begin{tabular}{c|ccc}
    \toprule[0.2em]
    Method & LPIPS$\downarrow$ & DISTS$\downarrow$ & PSNR$\uparrow$ \\
    \midrule
    DiffI2I$_{S2}$ (concat GT and input, Ours) & 0.2943 & 0.1038 & 27.52 \\
    DiffI2I$_{S2}$ (only GT) & 0.2951 & 0.1042 & 27.44 \\
    \bottomrule[0.2em]
    \end{tabular}%
  \label{tab:GT}%
\end{table}%

\begin{table}[t]
  \centering
  \caption{$4\times$ SR results on different resolutions (LPIPS$\downarrow$).}
    \begin{tabular}{c|ccc}
    \toprule[0.2em]
    Method & 256$\times$256 & 512$\times$512 & 1024$\times$1024 \\
    \midrule
    BybeGAN~\cite{BebyGAN} & 0.1210 & 0.1129 & 0.1088 \\
    SRDiff~\cite{srdiff} & 0.1469 & 0.1433 & 0.1412 \\
    DiffI2I$_{S2}$ (Ours) & 0.1088 & 0.1032 & 0.0992 \\
    \bottomrule[0.2em]
    \end{tabular}%
  \label{tab:resolution}%
\end{table}%

\section{Conclusion}
Traditional DMs have demonstrated impressive capabilities in image synthesis. However, when it comes to certain I2I tasks that demand strict adherence to GT constraints, a deviation from the traditional approach is necessary. The direct application of the DM framework, designed primarily for starting pixel-level generation from scratch, is inefficient and unsuitable for such I2I tasks. In this paper, we introduce a novel and efficient DM tailored specifically for I2I, named DiffI2I. This approach comprises three key components: CPEN, DI2Iformer, and a denoising network. The process begins with feeding the ground-truth image into CPEN$_{S1}$, resulting in the generation of a compact IPR. This IPR plays a crucial role in guiding the subsequent operations of the DI2Iformer. After that, the DM is trained to accurately estimate the compact IPR, as derived from CPEN$_{S1}$.
Remarkably distinguishing itself from traditional DMs, our DiffI2I demonstrates the remarkable ability to achieve significantly more precise estimations and effectively diminish artifacts in the generated results, all while employing far fewer iterations and a lighter denoising network. 
Furthermore, this reduction in the computation of DM facilitates the adoption of a joint optimization strategy encompassing CPEN$_{S2}$, DI2Iformer, and the denoising network. This approach is designed to mitigate the potential influence of estimation errors.
Our extensive experimental evaluations showcase DiffI2I's achievement of a new SOTA benchmark performance in the realm of I2I.

\bibliographystyle{IEEEtran}
\bibliography{egbib}

\end{document}